\documentclass[a4paper,fleqn]{cas-dc}

\usepackage[numbers,sort&compress]{natbib}
\usepackage[commandnameprefix=always]{changes}
\usepackage{subcaption}



\usepackage{tabularx} 
\usepackage{amsthm}
\usepackage{array}
\usepackage{algorithm}
\usepackage{algpseudocode}
\usepackage{booktabs}
\usepackage{threeparttable}
\usepackage[normalem]{ulem}
\theoremstyle{remark}          
\hypersetup{
  colorlinks=false,        
 pdfborder={0 0 1},       
            citebordercolor={0 1 0}, 
            linkbordercolor={1 0 0}, 
            urlbordercolor={0 1 1}   
}

\begin{document}
\let\WriteBookmarks\relax
\def\floatpagepagefraction{1}
\def\textpagefraction{.001}

\shorttitle{LocoMamba: Vision-Driven Locomotion via End-to-End Deep Reinforcement Learning with Mamba}

\shortauthors{Yinuo Wang and Xiaowen Tao}

\title [mode = title]{LocoMamba: Vision-Driven Locomotion via End-to-End Deep Reinforcement Learning with Mamba}                      


\author[1]{Yinuo Wang}[style=chinese]
\ead{allen.wang@woven.toyota}
\credit{Conceptualization, Methodology, Software, Validation, Investigation, Resources, Visualization, Writing- Original Draft, Writing- Review \& Editing}

\author[2]{Xiaowen Tao}[style=chinese]
\cormark[1]
\ead{taox@tcd.ie}
\credit{Conceptualization, Methodology, Formal analysis, Writing- Original Draft, Writing- Review \& Editing, Supervision}

\affiliation[1]{organization={Woven by Toyota},
    city={Palo Alto},
    state={CA},
    country={USA}}

\affiliation[2]{organization={School of Computer Science and Statistics, Trinity College Dublin},
    city={Dublin},
    country={Ireland}}

\cortext[cor1]{Corresponding author}

\begin{abstract}
Applying deep reinforcement learning (DRL) to quadrupedal robots is promising for obstacle negotiation, terrain locomotion, and real-world deployment. Existing methods either train blind agents that sacrifice foresight or adopt cross-modal fusion architectures with limitations; even Transformer-based models still suffer from the quadratic cost of self-attention. To overcome these challenges, we propose \emph{LocoMamba}, a vision-driven cross-modal DRL framework built on selective state-space models (Mamba). LocoMamba encodes proprioceptive and vision features into compact tokens, which are fused by stacked Mamba layers through near-linear-time selective scanning. This reduces latency and memory usage while preserving long-range dependencies. The policy is trained end-to-end with Proximal Policy Optimization under randomized terrain and obstacle-density curriculum, using a state-centric reward balancing task and safety. Evaluations in diverse static and dynamic obstacle settings show that LocoMamba significantly outperforms a Transformer baseline, achieving $48.9\%$ higher returns, $30.4\%$ longer distance, and $48.9\%$ fewer collisions on trained terrains, while exhibiting stronger generalization to unseen scenarios. Moreover, it converges faster with limited compute, enhancing training efficiency. Overall, LocoMamba improves locomotion performance and reduces training cost, supporting rapid iteration and deployment of quadruped learning algorithms. A repository is hosted at \url{https://github.com/allen-quad-robot/locomamba}.

\end{abstract}


\begin{highlights}
\item \textbf{State-of-the-art LocoMamba:} To the best of our knowledge, this is the first vision-driven cross-modal DRL framework for quadrupedal locomotion that utilizes the selective state-space model Mamba as the fusion backbone, enabling foresightful and efficient control.

\item \textbf{Effective proprioception and depth encoders:} A compact input representation is introduced in which an MLP embeds proprioceptive states and a CNN patchifies depth images into spatial tokens tailored for Mamba-based fusion. This design provides immediate state estimates and look-ahead while reducing sensitivity to appearance variation, thereby improving computational efficiency and training stability.

\item \textbf{Efficient cross-modal Mamba fusion backbone:} Encoded tokens from proprioception and depth are fused using stacked Mamba layers via selective state-space scanning, achieving near-linear time and memory scaling. The backbone supports long-horizon modeling, remains robust to token length and image resolution, and provides a regularizing inductive bias through input-gated, exponentially decaying dynamics.

\item \textbf{Robust end-to-end RL training scheme:} The policy is trained with PPO using Mamba-fused cross-modal features, complemented by terrain and appearance randomization and an obstacle-density curriculum. A compact, state-centric reward balances task-aligned progress, energy efficiency, and safety, enabling stable learning and consistent performance.

\item \textbf{Comprehensive evaluation:} Extensive experiments are conducted with static and moving obstacles and uneven terrain, and demonstrate consistent gains over state-of-the-art (SOTA) in terms of return, collision times, and distance moved, along with faster convergence under the same compute budget.
\end{highlights}

\begin{keywords}
Quadrupedal Robots\sep Vision-driven Locomotion \sep Reinforcement Learning\sep Cross-modal Fusion\sep End-to-end Policy Learning
\end{keywords}

\maketitle

\section{Introduction}
\label{sec:first}
Quadrupedal robots provide mobility in environments where wheeled platforms are ineffective, such as stairs, rubble, soft or deformable substrates, and cluttered indoor or outdoor settings, enabling applications in inspection, disaster response, agriculture, and planetary exploration \citep{fan2024review}. Robust locomotion control is therefore a foundational capability for practical quadrupedal systems, underpinning safe navigation, dependable mission execution, and reliable operation across diverse terrains and disturbance conditions \citep{carpentier2021recent}.

Current approaches to quadrupedal locomotion control fall into two broad categories. Classical model-based pipelines depend on accurate dynamics and terrain models, explicit contact planning, and extensive parameter tuning, which constrain scalability and complicate deployment \citep{xin2021robust}. Learning-based methods mitigate this burden and enable end-to-end policies that couple perception with control \citep{ha2025learning}. Within this paradigm, deep reinforcement learning has emerged as the dominant approach because it optimizes task-driven closed-loop policies through interaction with the environment and produces robust, adaptive behaviors \citep{zhang2022deepreinforcementlearningforreal}.

DRL methods have substantially advanced quadrupedal locomotion, enabling traversal of uneven terrain \citep{xie2021dynamics} and operation under challenging conditions such as mud, snow, and running water \citep{lee2020learning}. However, most studies continue to train blind controllers that rely only on proprioception and achieve stability primarily through large-scale simulation and domain randomization. Despite their robustness, blind agents lack foresight because they receive no exteroceptive input, so they react only after contact and struggle to avoid obstacles proactively or to plan foot placement on irregular ground \citep{xie2022glide}.

On the other hand, cross-modal fusion with visual input has gained traction. Vision complements proprioception by providing look-ahead, enabling early perception of distant obstacles, anticipation of terrain changes before contact, and timely trajectory adjustments that reduce collisions and improve foot placement \citep{han2025multimodal}. Nevertheless, current fusion architectures exhibit notable trade-offs. Recurrent models often suffer from vanishing gradients and limited capacity for long-horizon dependencies \citep{li2024rapid}. Hierarchical designs complicate optimization and propagate errors across levels \citep{jain2019hierarchical}. Transformer-based fusion, although expressive, incurs quadratic memory use and computational cost with the number of tokens, which constrains sequence length, spatial resolution, and training efficiency \citep{yang2021learning}.

In this paper, LocoMamba, a vision-driven end-to-end DRL framework that uses Mamba as a selective state-space model (SSM) backbone, is introduced to enable efficient quadrupedal locomotion. A lightweight multilayer perception (MLP) embeds proprioceptive states to provide accurate estimates of the robot state for immediate reaction, while a compact convolutional neural network (CNN) patchifies depth images to supply look-ahead for negotiating uneven terrain and large obstacles. Stacked Mamba layers fuse tokens through selective state-space scanning with near-linear time complexity, using recurrent updates whose cost scales with sequence length and thereby lowering latency and memory use compared with quadratic self-attention. The streaming formulation accommodates variable-length inputs, enabling longer visual context and higher token resolution, while input-gated, exponentially decaying dynamics provide a regularizing inductive bias that mitigates overfitting.
Policies are trained with Proximal Policy Optimization (PPO) under terrain and appearance randomization together with an obstacle-density curriculum, which broadens environmental coverage and progressively increases difficulty to stabilize on-policy learning. The reward is compact and state-centric, encouraging task-aligned progress, promoting energy-efficient actuation, and enforcing safety via penalties for collisions and falls.

\textbf{The main contributions of this paper are summarized as follows:}
\begin{enumerate}[1)]
\item \textbf{State-of-the-art LocoMamba:} To the best of our knowledge, this is the first vision-driven cross-modal DRL framework for quadrupedal locomotion that utilizes the selective state-space model Mamba as the fusion backbone, enabling foresightful and efficient control.

\item \textbf{Effective proprioception and depth encoders:} A compact input representation is proposed in which an MLP embeds proprioceptive states and a CNN patchifies depth images into spatial tokens tailored for Mamba-based fusion. This design provides immediate state estimates and foresight while reducing sensitivity to appearance variation, thereby improving computational efficiency and training stability.

\item \textbf{Efficient cross-modal Mamba fusion backbone:} The encoded tokens are fused from proprioception states and depth images using stacked Mamba layers via selective state-space scanning, achieving near-linear time and memory scaling. The backbone supports long-horizon modeling, remains robust to token length and image resolution, and provides a regularizing inductive bias through input-gated, exponentially decaying dynamics.

\item \textbf{Robust end-to-end RL training scheme:} The policy is trained with PPO using Mamba-fused cross-modal features, complemented by terrain and appearance randomization and an obstacle-density curriculum. A compact, state-centric reward balances task-aligned progress, energy efficiency, and safety, enabling stable learning and consistent performance.

\item \textbf{Comprehensive evaluation:} Extensive experiments are conducted with static and moving obstacles and uneven terrain, and demonstrate consistent gains over state-of-the-art (SOTA) in terms of return, collision times, and distance moved, along with faster convergence under the same compute budget.
\end{enumerate}

The remainder of this paper is organized as follows. Section \ref{sec:second} reviews related work on blind and cross-modal quadruped locomotion. Section \ref{sec:third} presents the LocoMamba methodology, including the problem formulation, proprioception and depth encoders, and the Mamba fusion backbone with training objectives. Section \ref{sec:fourth} describes the experimental setup and evaluation protocol. Section \ref{sec:fifth} presents the experimental evaluation, covering the main results and ablations, as well as extended validation of efficiency and generalization. Finally, Section \ref{sec:sixth} concludes the paper and outlines directions for future research.

\section{Related Work}
\label{sec:second}
\subsection{Learning-Based Quadrupedal Locomotion}
{Research on quadrupedal locomotion has evolved along two principal paradigms: model-based optimal control and learning-based approaches. Early model-based optimal control relied on template dynamics and heuristic regulation of balance, leg compliance, and foot placement, establishing fundamental principles for dynamic locomotion \citep{miura1984dynamic,liu2014planning,habu2018simple,bledt2018cheetah}. Subsequent model-based optimal control systems formalized locomotion as constrained optimal control, including trajectory optimization over centroidal or full-body dynamics and model predictive control with explicit contact constraints and task hierarchies \citep{grandia2019feedback,yao2021hierarchy,amatucci2024accelerating,elobaid2025adaptive}.} For example, \citep{di2018dynamic} developed real-time MPC and task-space whole-body control on torque-controlled quadrupeds, demonstrating stable trotting, stair traversal, and robust disturbance rejection. \citep{ding2019real} refined MPC cost design and contact modeling to improve tracking and compliance under high-frequency control budgets. \citep{carius2019trajectory} studied trajectory optimization and contact-aware planning for foothold selection and force allocation, improving reliability on structured terrain. However, these approaches require accurate models of the environment and system dynamics and substantial manual tuning, thereby limiting their applicability in complex and variable settings.

On the other hand, model-free RL can learn general policies for challenging conditions \citep{li2021reinforcement,margolis2024rapid,bussola2025guided,ha2025learning}. For example, \citep{hwangbo2019learning} trained torque-level policies with dynamics randomization and privileged learning to achieve fast, robust trotting on uneven terrain. \citep{tan2018sim} used large-scale domain randomization and dynamics perturbations to transfer simulation-trained policies to real quadrupeds with strong disturbance rejection. \citep{kumar2021rma} introduced rapid motor adaptation that augments a proprioceptive policy with an online adaptation module, enabling quick recovery under payload shifts and terrain changes. However, most approaches are proprioception-only and thus blind to exteroceptive cues, which limits foresight for obstacle negotiation and precise foothold planning. In this work, both the visual and proprioceptive inputs are fused to obtain a richer state representation, enabling the policy to anticipate terrain changes and plan trajectories online while maintaining continuous motion.

\subsection{Vision-Driven RL for Quadrupedal Locomotion}
To extend RL beyond state-only inputs, a growing body of work leveraged visual observations for locomotion control \citep{han2025multimodal}. For example, \citet{yu2021visual} trained a vision-based controller that incorporated exteroceptive input directly into the RL loop and achieved traversal of uneven terrain and complex obstacles. \citet{duan2024learning} learned a controller in simulation with heightmap observations and then trained a depth-to-heightmap predictor from depth and state histories, enabling vision-guided locomotion with transfer to hardware on challenging terrains. Complementing these efforts, \citet{fahmi2022vital} proposed the ViTAL framework, which used vision for terrain-aware planning by decomposing decision making into foothold selection and pose adaptation, and improved safety and reliability on irregular surfaces. Hierarchical formulations were also explored; for example, \citet{jain2019hierarchical} employed hierarchical RL in which high-level visual policies guided low-level motor control with sensor inputs.

Despite these advances, common fusion strategies involved trade-offs. Simple concatenation underused spatial structure, and hierarchical decompositions increased optimization complexity and risked error propagation across levels \citep{singh2022reinforcement}. In this work, complex hierarchical designs are avoided and a simple end-to-end schema is adopted that fuses proprioceptive and depth inputs within a single policy trained with PPO, enabling visual look-ahead while maintaining a streamlined learning pipeline.

\subsection{State-Space Models and Cross-Modal RL}
Cross-modal RL has attracted significant attention. Typical fusion backbones include recurrent neural networks (RNNs) and Transformers. For example, \citep{xiao2024egocentric} designed an egocentric visuomotor RL framework that fused egocentric depth and proprioception with long short-term memory (LSTM) units, enabling a quadruped to negotiate obstacles and traverse cluttered scenes in the real world. \citep{lai2024world} proposed a gated recurrent unit (GRU)–based world-model formulation that fuses depth images and proprioceptive readings into a recurrent latent state for visual legged locomotion, improving control under partial observability. \citep{yang2021learning} introduced a cross-modal Transformer policy that ingests depth tokens and proprioceptive features end-to-end, achieving stronger terrain anticipation and sim-to-real transfer than blind baselines; subsequent work scaled such tokenized proprioception-vision Transformer policies to heterogeneous robot datasets to improve generalization. However, RNNs often suffer from vanishing gradients and limited long-horizon capacity, which complicates optimization in extended sequences. Transformer models offer strong expressivity but incur quadratic memory and computation with the number of tokens, which constrains sequence length, spatial resolution, and training efficiency.

SSMs \citep{kalman1960new} offer an alternative by updating a compact recurrent state per token and scaling near-linearly with sequence length. Building on the S4 model \citep{gu2021efficiently}, which demonstrated efficient long-range dependency modeling on long-sequence benchmarks, \citep{gu2023mamba} introduced Mamba , a selective SSM whose input-dependent parameters and hardware-aware parallel scan yield linear-time sequence modeling with favorable throughput and memory characteristics across modalities.

Mamba-style SSMs have also been explored for multimodal fusion. AlignMamba augments a multimodal Mamba backbone with token-level optimal-transport alignment and a global distributional alignment objective to improve cross-modal consistency \citep{li2025alignmamba}. COMO (Cross-Mamba Interaction with Offset-Guided Fusion) addresses sensor misalignment for multimodal object detection while retaining the efficiency benefits of selective scanning \citep{liu2026cross}. AV-Mamba applies selective SSMs to audio-visual question answering and reports improvements over Transformer baselines \citep{huang2024av}. FusionMamba \citep{xie2024fusionmamba} and DepMamba \citep{ye2025depmamba} further demonstrate SSM-based gains in multimodal image fusion and audio-visual affect analysis. {Unlike prior cross-modal Mamba variants that focus on perception-level multimodal alignment, our work investigates its benefits for robotic locomotion control, where real-time decision latency, causal sensor fusion, and long-horizon dependency are critical. This setting highlights Mamba’s suitability for efficient and temporally consistent proprioception–vision fusion within an end-to-end RL framework.} To the best of our knowledge, this is the first work to use cross-modal Mamba for quadrupedal locomotion.

\section{Methodology}
\label{sec:third}
\begin{figure*}
\centering
\includegraphics[width=1.0\linewidth]{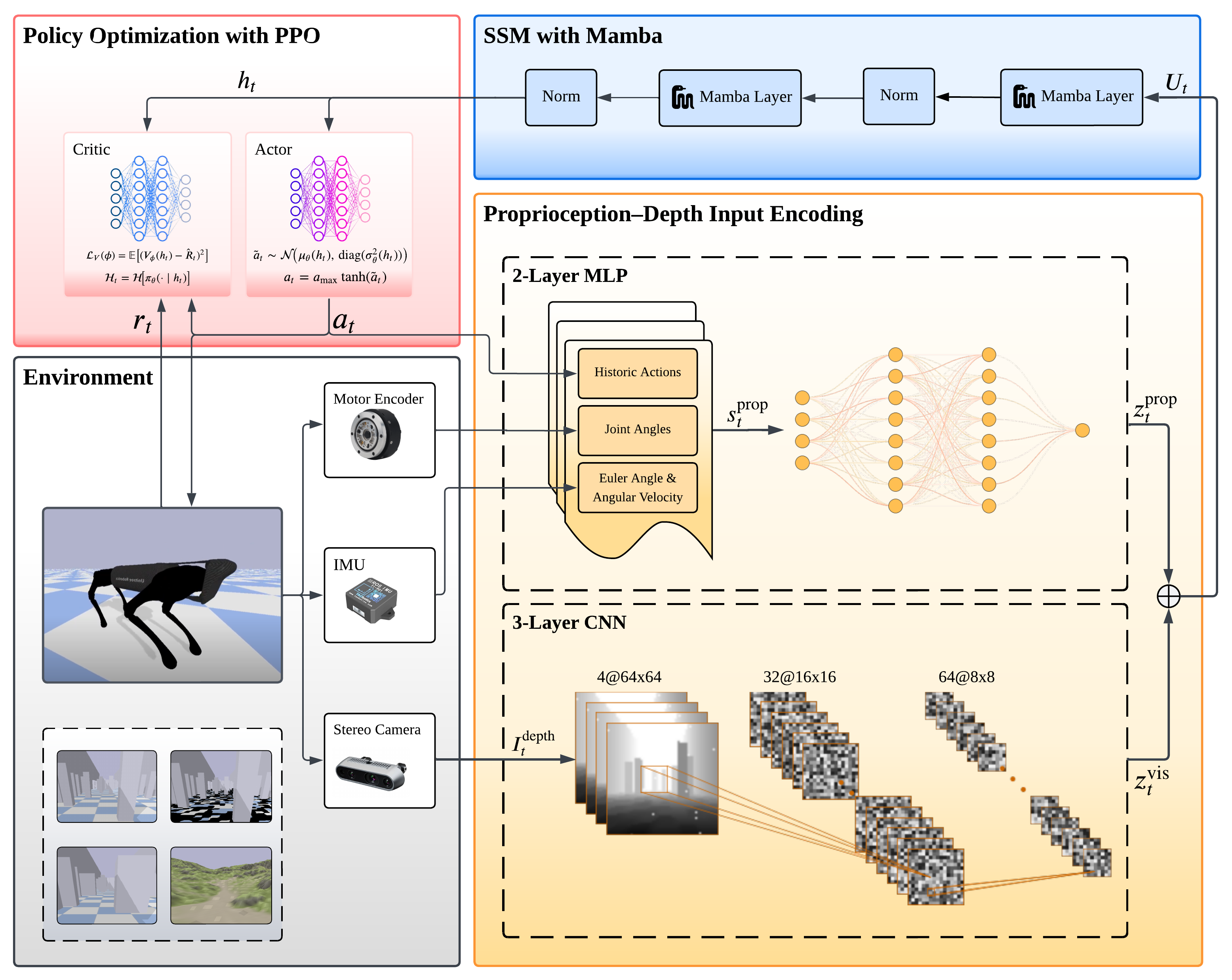}
\caption{{Overall architecture of LocoMamba. Proprioception and depth are encoded into tokens by MLP/CNN, fused by a Mamba SSM backbone, and optimized with PPO.}}
\label{fig:overall_archi}
\end{figure*}

Fig. \ref{fig:overall_archi} presents the overall architecture of LocoMamba. The pipeline comprises three components trained end to end. First, proprioceptive states and depth images are encoded into a compact tokenized latent space: an MLP maps the proprioceptive vector to a state token, and a lightweight CNN patchifies the depth image into spatial tokens. Second, the concatenated token stream is fused by stacked Mamba SSM layers via selective state-space scanning, which updates a compact recurrent state per token, yields near linear time and memory scaling, accommodates variable token counts and resolutions, and preserves long-horizon temporal context. Third, policy and value heads consume the fused features and are optimized with PPO \citep{schulman2017proximal} under a compact state-centric reward that balances task-aligned progress, energy efficiency, and safety.

\subsection{Proprioception--Depth Input Encoding}

As described in Section~\ref{sec:first}, an agent is considered to uses both proprioceptive state and depth information for decision making. At time $t$, the observation is
\begin{equation}
o_t = \{\, s^{\text{prop}}_t,\; I^{\text{depth}}_t \,\},
\end{equation}
where $s^{\text{prop}}_t \in \mathbb{R}^{D_p}$ denotes the proprioceptive vector and $I^{\text{depth}}_t \in \mathbb{R}^{H \times W}$ denotes a first-person-view depth image that captures obstacles and terrain ahead of the robot. Temporal aggregation of the proprioceptive vector and the four most recent depth frames is handled by the fusion backbone described in Section~\ref{sec:mamba}.

\paragraph{Encoders and Tokenization.}
Each modality is encoded with a lightweight, domain-specific network and unify them in a shared latent space. The proprioceptive vector is mapped by an MLP to a compact token:
\begin{equation}
z^{\text{prop}}_t = f_{\text{MLP}}(s^{\text{prop}}_t), \quad z^{\text{prop}}_t \in \mathbb{R}^{d_p}.
\end{equation}
The depth image is processed by a compact CNN \citep{li2021survey} to produce a sequence of spatial tokens:
\begin{equation}
z^{\text{vis}}_t = f_{\text{CNN}}(I^{\text{depth}}_t), \quad z^{\text{vis}}_t \in \mathbb{R}^{N \times d_v}.
\end{equation}
With patch size $P$, the number of spatial tokens is
\begin{equation}
N = \frac{H}{P} \cdot \frac{W}{P}.
\end{equation}

To form a single cross-modal stream, both embeddings are projected to a common width $d$. The projected proprioceptive token is
\begin{equation}
\tilde{z}^{\text{prop}}_t = W_p \, z^{\text{prop}}_t, \quad \tilde{z}^{\text{prop}}_t \in \mathbb{R}^{d}.
\end{equation}
The projected visual tokens are
\begin{equation}
\tilde{z}^{\text{vis}}_t = W_v \, z^{\text{vis}}_t, \quad \tilde{z}^{\text{vis}}_t \in \mathbb{R}^{N \times d}.
\end{equation}
where $W_p$ and $W_v$ are learned projections, and $N$ denotes the number of spatial tokens.
Then the modalities are concatenated into one token sequence
\begin{equation}
U_t = \big[ \tilde{z}^{\text{prop}}_t \,;\, \tilde{z}^{\text{vis}}_t \big], \quad U_t \in \mathbb{R}^{(1+N) \times d}.
\end{equation}
Finally, learned position indices and modality tags are added, followed by per-token layer normalization, to obtain
\begin{equation}
\hat{U}_t = \mathrm{LN}\!\left( U_t + E_{\text{pos}}^{\text{spat}} + E_{\text{mod}} \right),
\end{equation}
where $E_{\text{pos}}^{\text{spat}} \in \mathbb{R}^{(1+N)\times d}$ encodes spatial indices (the proprioceptive token receives a distinct index), and $E_{\text{mod}} \in \mathbb{R}^{(1+N)\times d}$ distinguishes proprioceptive and visual tokens.

\subsection{State Space Modelling with Mamba}
\label{sec:mamba}

\begin{figure*}
\centering
\includegraphics[width=0.6\linewidth,angle=90]{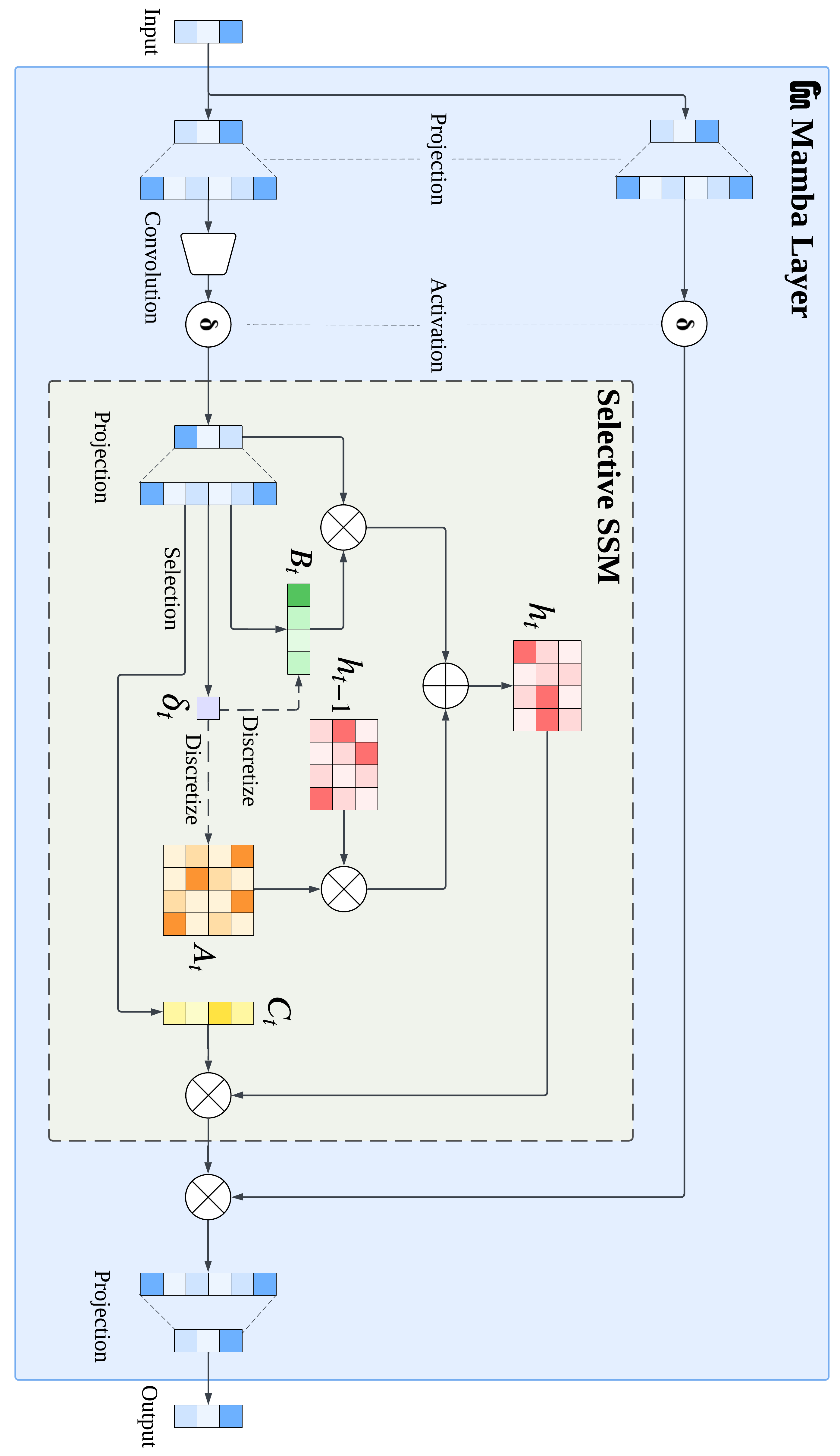}
\caption{Schematic of a Mamba SSM Layer.}
\label{fig:mamba}
\end{figure*}

The cross-modal token sequence $\hat{U}_t$ is fused with a stack of Mamba SSM layers \citep{gu2023mamba} (Fig. \ref{fig:mamba}). Each layer scans the token stream with input-dependent state updates and carries a compact recurrent state across tokens and time steps, enabling efficient long-horizon modeling.

\paragraph{Per-Layer Selective SSM.}
Let $u_{t,k} \in \mathbb{R}^{d}$ denote the $k$-th token at time $t$ from $\hat{U}_t$ ($k{=}1$ is the proprioceptive token, $k{=}2\ldots 1{+}N$ are visual tokens in raster order). A Mamba layer maintains a hidden state $x_{t,k} \in \mathbb{R}^{h}$ and computes
\begin{equation}
\label{eq:ssm-update}
x_{t,k+1} \;=\; \bar{A}_{t,k}(u_{t,k})\, x_{t,k} \;+\; \bar{B}_{t,k}(u_{t,k})\, u_{t,k},
\end{equation}
\vspace{-2\baselineskip}
\begin{equation}
\label{eq:ssm-output}
y_{t,k} \;=\; \bar{C}_{t,k}(u_{t,k})\, x_{t,k} \;+\; \bar{D}_{t,k}(u_{t,k})\, u_{t,k},
\end{equation}
where $\bar{A}_{t,k},\bar{B}_{t,k},\bar{C}_{t,k},\bar{D}_{t,k}$ are input-gated, token-dependent parameters produced by lightweight affine transforms of $u_{t,k}$. The scan proceeds over $k{=}1\ldots (1{+}N)$, and the initial state at each time step is carried from the previous step, $x_{t,1} \leftarrow x_{t-1,\,1+N}$. The output tokens of the layer are $Y_t=\{y_{t,k}\}_{k=1}^{1+N} \in \mathbb{R}^{(1+N)\times d}$. The token width $h$ and the hidden state dimension of the SSM $d$ are set to equal by a learned projection when needed. {In this work, a two-layer stacked Mamba backbone is adopted as a balanced configuration that captures both short- and mid-term temporal dependencies while maintaining training stability and computational efficiency.}

\paragraph{Stacked Fusion Backbone and Complexity.}
The $L_{\text{m}}$ Mamba layers are stacked with residual connections and layer normalization:
\begin{equation}
H^{(\ell+1)}_t \;=\; \mathrm{LN}\!\left( Y^{(\ell)}_t + H^{(\ell)}_t \right), \qquad \ell = 0,\ldots,L_{\text{m}}-1,
\end{equation}
with $H^{(0)}_t \!=\! \hat{U}_t$ and $Y^{(\ell)}_t$ computed by \eqref{eq:ssm-update}–\eqref{eq:ssm-output} using inputs $H^{(\ell)}_t$. The selective SSM update yields near-linear time and memory in the token count $(1{+}N)$, in contrast to the quadratic cost of global self-attention \citep{singh2022reinforcement}, and naturally supports streaming over time by reusing the carried state $x_{t,1}$.

\paragraph{Projection Head for Policy Inputs.}
To obtain a compact feature for control, modality-aware pooling is performed on the final layer outputs $H^{(L_{\text{m}})}_t$. Let $y^{\text{prop}}_t$ be the proprioceptive token and $\{ y^{\text{vis}}_{t,i} \}_{i=1}^{N}$ be the visual tokens from $H^{(L_{\text{m}})}_t$. Then they can be calculated as:
\begin{equation}
\bar{y}^{\text{vis}}_t = \frac{1}{N}\sum_{i=1}^{N} y^{\text{vis}}_{t,i}, \qquad
h_t = f_{\text{head}}\!\left( \big[\, y^{\text{prop}}_t \,;\, \bar{y}^{\text{vis}}_t \,\big] \right) \in \mathbb{R}^{d_h},
\end{equation}
where $f_{\text{head}}$ is a small MLP. The fused feature $h_t$ parameterizes the policy and value functions in Section~\ref{sec:ppo}.

\paragraph{Positional Coding and Robustness.}
Spatial indices are encoded in $E_{\text{pos}}^{\text{spat}}$ and preserved throughout the scan. When temporal indices are used, an additional code $E_{\text{pos}}^{\text{time}}$ is added to the tokens before the first Mamba layer. The recurrent state $x_{t,k}$ provides a causal, exponentially decaying memory that promotes temporally consistent features and improves robustness to variable token counts and resolutions, while keeping latency and memory use low.

\subsection{Policy Optimization with PPO}
\label{sec:ppo}

\paragraph{Markov Decision Process (MDP).}
The locomotion can be modeled as a discounted MDP
$\mathcal{M}=(\mathcal{S},\mathcal{A},P,r,\gamma)$ with discount $\gamma\in(0,1)$\citep{richard1957mark}.
At time $t$, the agent receives observation $o_t$ from environment and obtains a fused feature
$h_t=f_{\text{fuse}}(o_t)\in\mathbb{R}^{d_h}$ from the Mamba backbone
(Sections \ref{sec:mamba}). The objective is to maximize the expected discounted return
\begin{equation}
J(\theta)=\mathbb{E}\!\left[\sum_{t=0}^{T-1}\gamma^t r_t\right],
\end{equation}
where the expectation is over on-policy trajectories, $T$ is the horizon, and $r_t=r(s_t,a_t)$.

The action is also a 12-dimensional vector that controls the change of all the joint
angles. A Gaussian policy is used in
the unconstrained space with squashing to actuator limits:
\begin{equation}
\tilde{a}_t \sim \mathcal{N}\!\left(\mu_\theta(h_t),\, \operatorname{diag}\!\big(\sigma_\theta^2(h_t)\big)\right).
\end{equation}
\vspace{-2\baselineskip}
\begin{equation}
a_t = a_{\max}\,\tanh(\tilde{a}_t).
\end{equation}

where $\mu_\theta(\cdot)$ and $\sigma_\theta(\cdot)$ are outputs of the policy head.

\paragraph{Policy and Value Parameterization.}
Two MLP heads consume \(h_t\): the first produces \(\mu_\theta(h_t)\) and \(\log\sigma_\theta(h_t)\), which parameterize the policy \(\pi_\theta(a_t \mid h_t)\), and the second outputs \(V_\phi(h_t)\) as the state-value estimate.

\paragraph{PPO Objective and Estimation.}
The policy is optimized with PPO \citep{schulman2017proximal} using the clipped surrogate. With likelihood ratio
\begin{equation}
\rho_t(\theta)=\frac{\pi_\theta(a_t\mid h_t)}{\pi_{\theta_{\text{old}}}(a_t\mid h_t)},
\end{equation}
the policy loss is
\begin{equation}
\mathcal{L}_{\text{clip}}(\theta)=
\mathbb{E}\Big[\min\big(\rho_t(\theta)A_t,\ \operatorname{clip}(\rho_t(\theta),1-\epsilon,1+\epsilon)\,A_t\big)\Big],
\end{equation}
where $\epsilon>0$ is the clip parameter and $A_t$ are advantages.
Advantages use generalized advantage estimation (GAE),
\begin{equation}
\delta_t=r_t+\gamma V_\phi(h_{t+1})-V_\phi(h_t),\qquad
A_t=\sum_{l=0}^{T-1-t}(\gamma\lambda)^l\,\delta_{t+l},
\end{equation}
where $\lambda\in[0,1]$ controls bias–variance trade-offs. The critic and entropy terms are
\begin{equation}
\mathcal{L}_V(\phi)=\mathbb{E}\big[(V_\phi(h_t)-\hat{R}_t)^2\big],\qquad
\mathcal{H}_t=\mathcal{H}\!\big[\pi_\theta(\cdot\mid h_t)\big],
\end{equation}
where $\hat{R}_t$ is the empirical return and $\mathcal{H}$ denotes entropy.
The total loss is
\begin{equation}
\mathcal{J}(\theta,\phi)= -\,\mathcal{L}_{\text{clip}}(\theta)
+ \beta_V\,\mathcal{L}_V(\phi)
- \beta_H\,\mathbb{E}[\mathcal{H}_t],
\end{equation}
where $\beta_V,\beta_H\ge0$ weight value and entropy terms.

\section{Implementation}
\label{sec:fourth}
\subsection{Environment Setup}
All experiments are conducted on a laptop equipped with an Intel Core Ultra 7 155H CPU (22 cores, 1.4\,GHz base clock) and an NVIDIA GeForce RTX~500 Ada GPU (4\,GB, CUDA-12.9). The operating system is Ubuntu~22.04. The physics simulation is run with PyBullet \citep{coumans2021pybullet} and all models are implemented in Python 3.8 with PyTorch~2.4.1.

The model is evaluated across three simulated environments that vary in terrain difficulty and obstacle dynamics:
\begin{itemize}
  \item \textbf{Thin Obstacle}: flat terrain populated with numerous thin cuboid obstacles.
  \item \textbf{Rugged Terrain}: uneven, discontinuous ground with a maximum height of 5 cm, requiring careful placement of foothold.
  \item \textbf{Dynamic Obstacle}: thin obstacles that move in random directions.
\end{itemize}

\begin{figure}[t]
\centering
\begin{subfigure}{\linewidth}
\centering
\includegraphics[width=0.49\linewidth]
{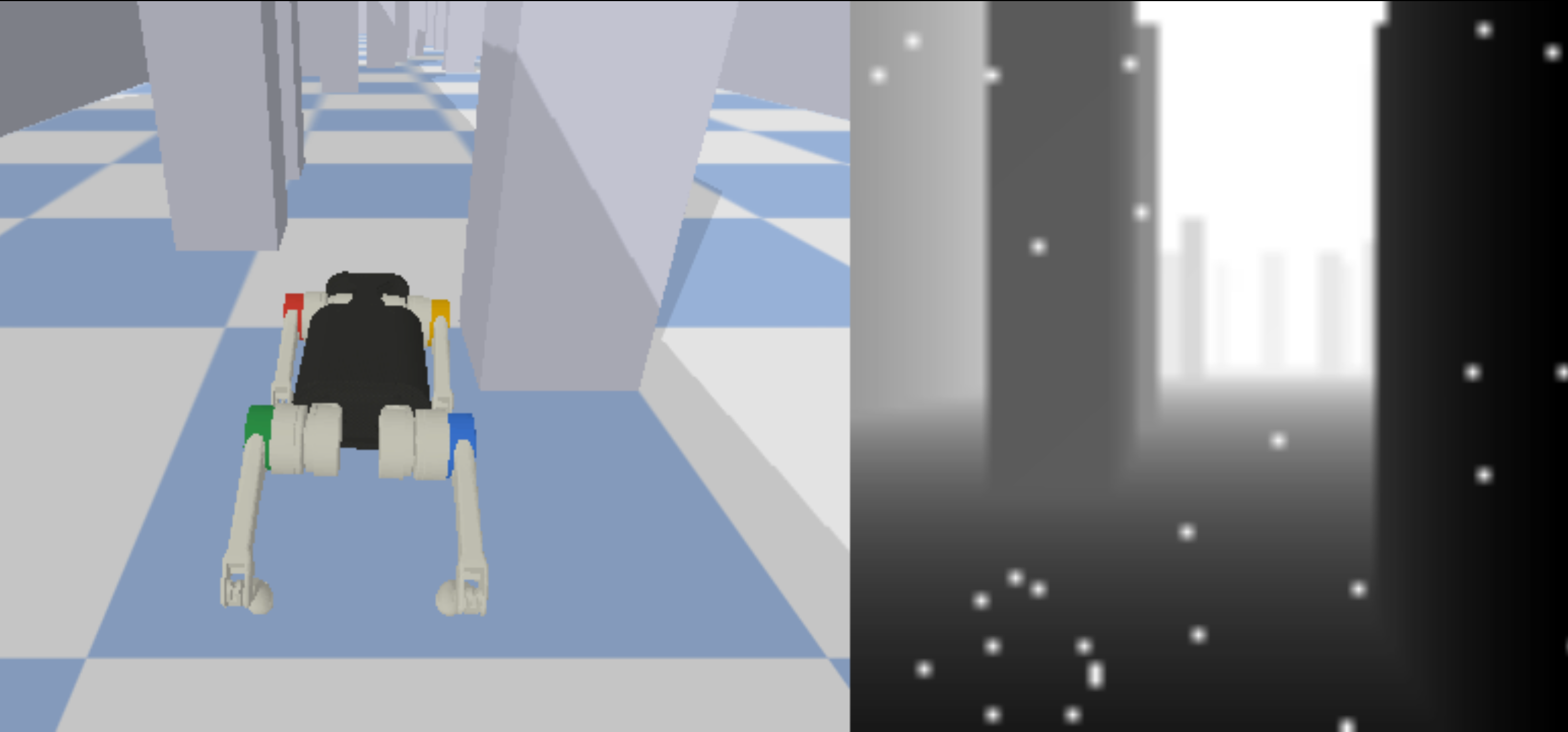}
\hfill
\includegraphics[width=0.49\linewidth]{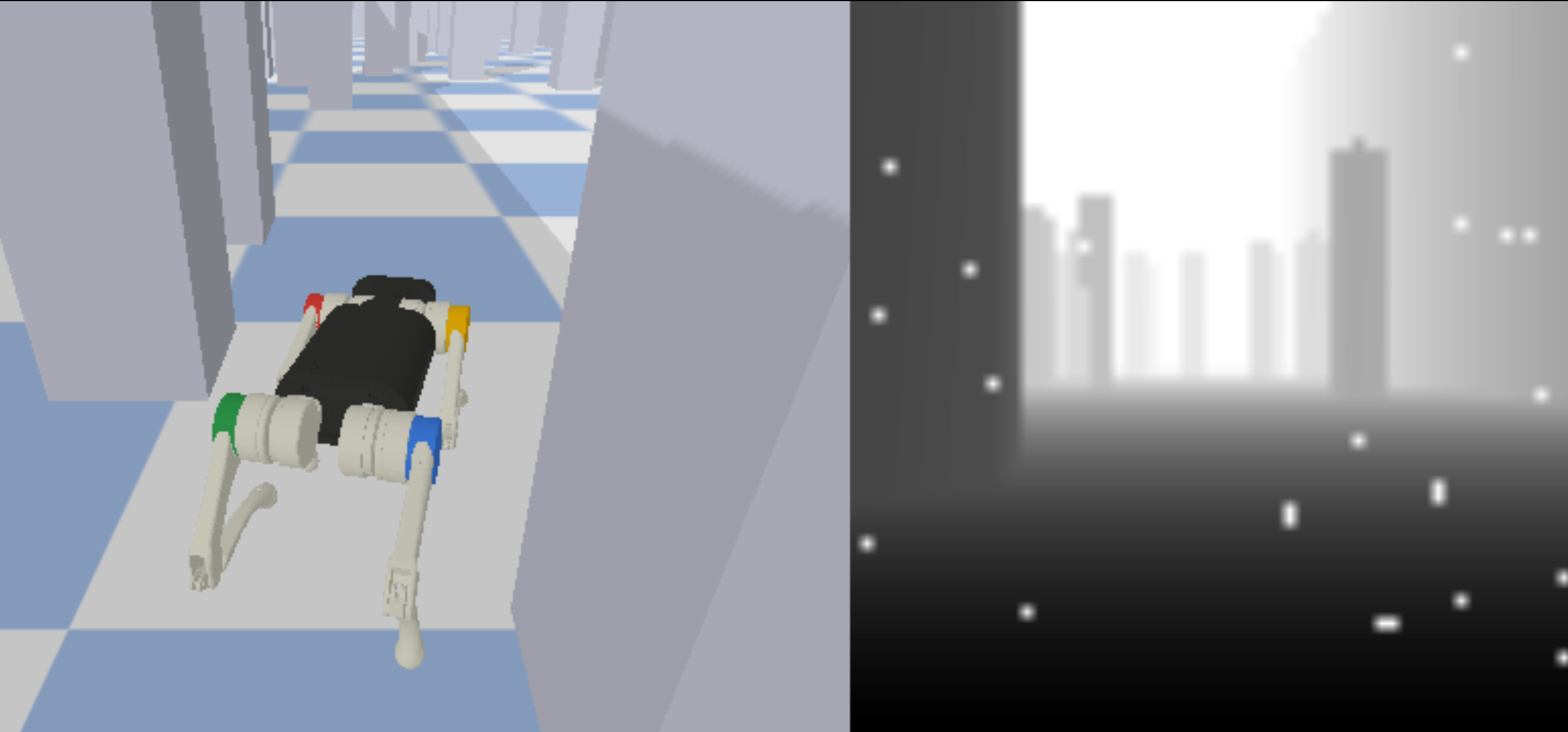} 
    \caption{}
    \label{fig:suba}
\end{subfigure}
\begin{subfigure}{\linewidth}
\centering
\includegraphics[width=0.49\linewidth]
{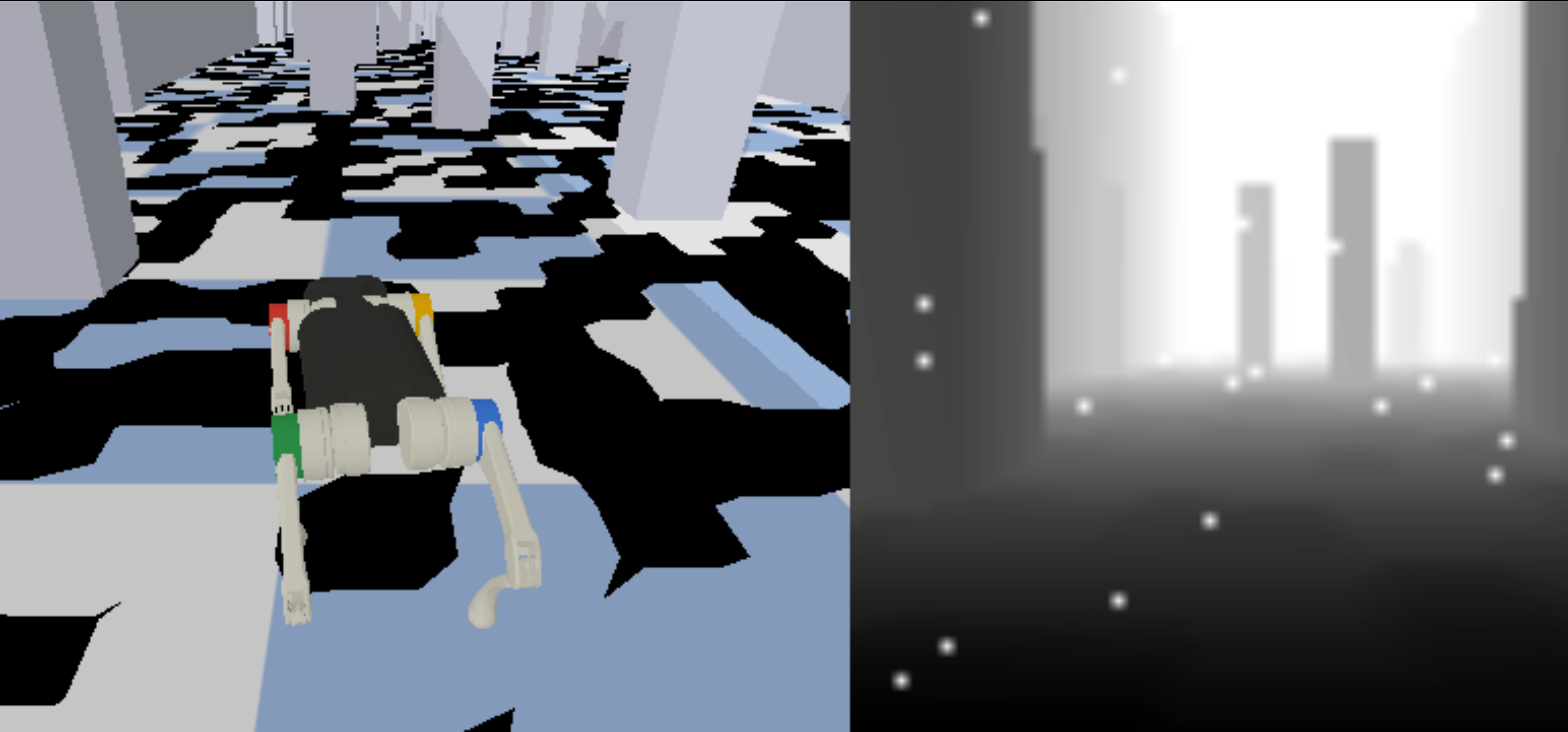}
\hfill
\includegraphics[width=0.49\linewidth]{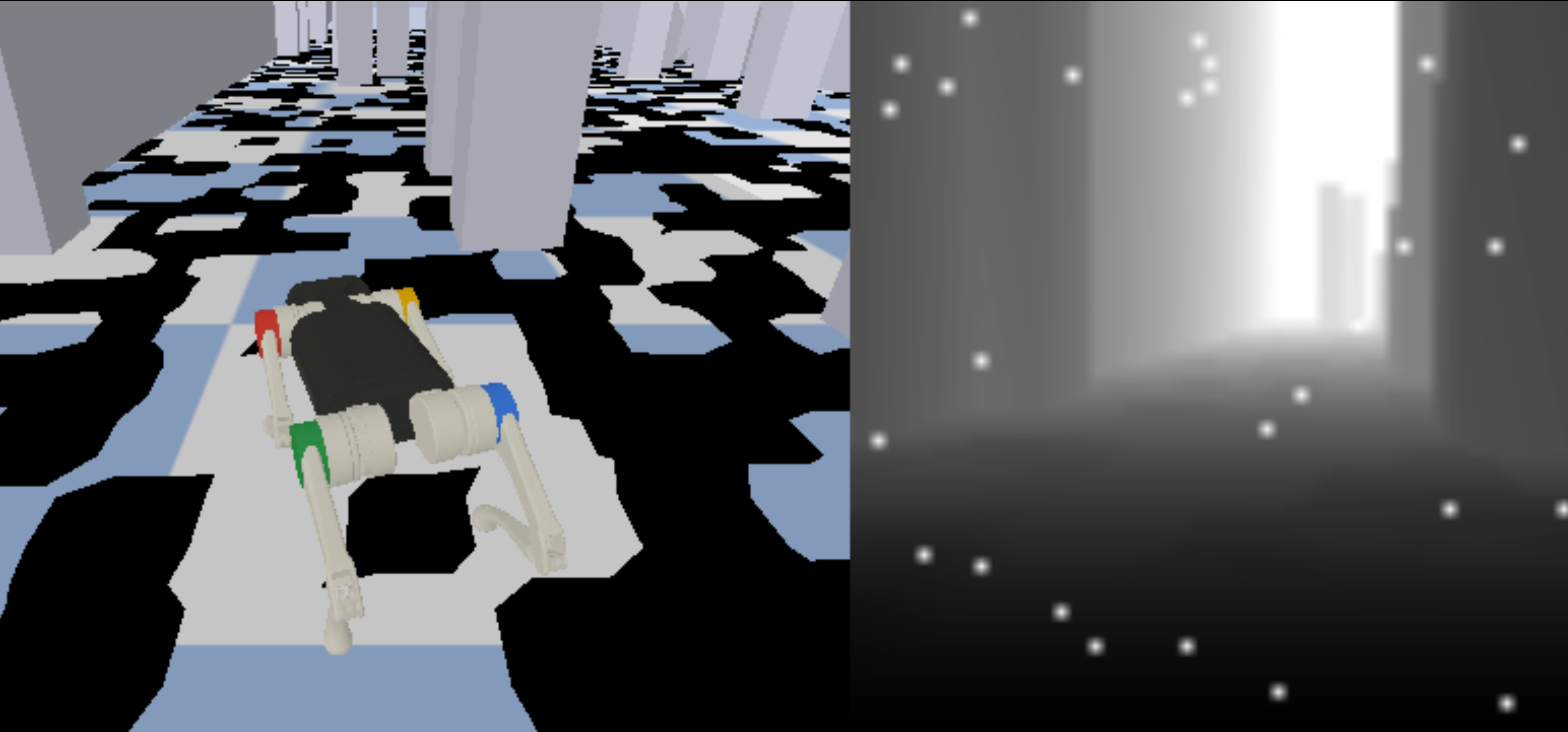} 
    \caption{}
    \label{fig:subb}
\end{subfigure}
\begin{subfigure}{\linewidth}
\centering
\includegraphics[width=0.49\linewidth]
{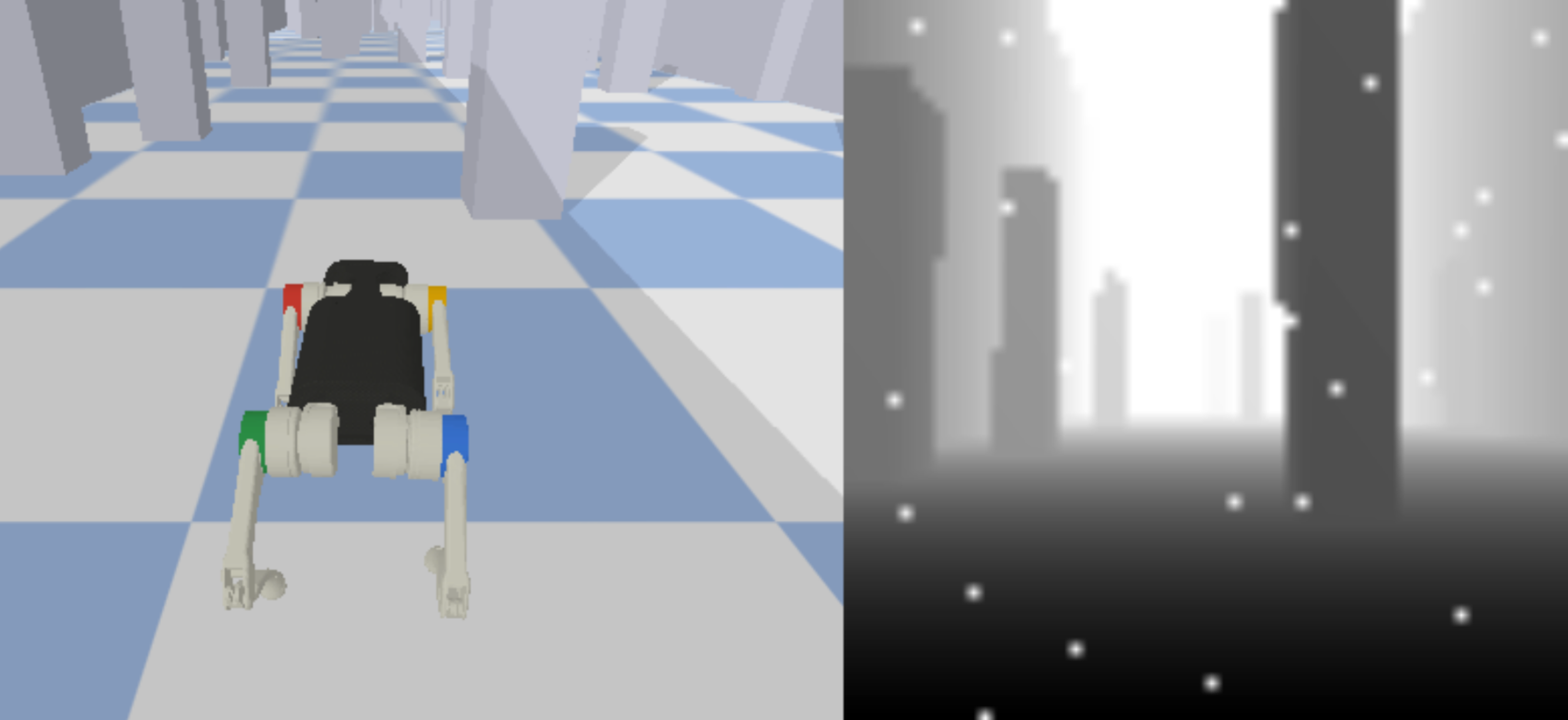}
\hfill
\includegraphics[width=0.49\linewidth]{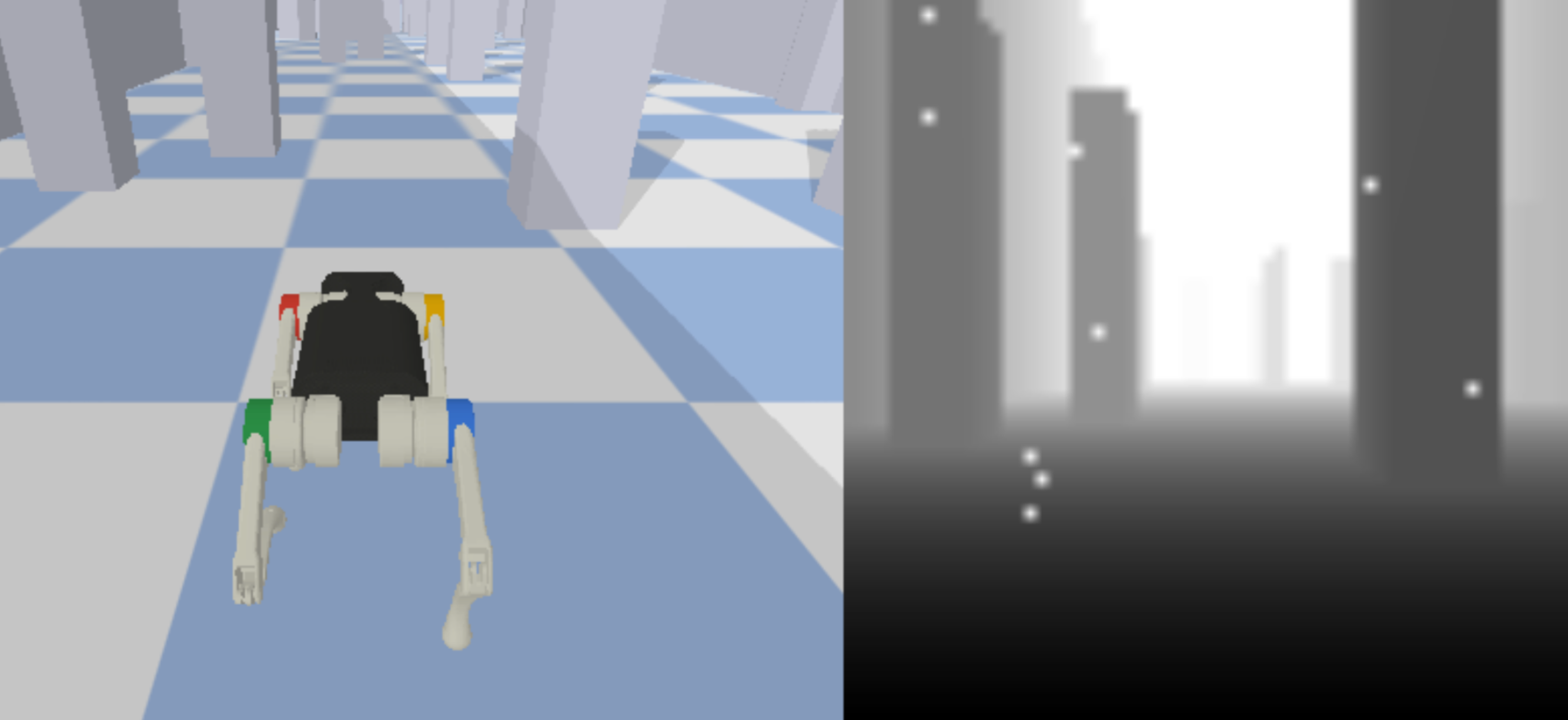} 
    \caption{}
    \label{fig:subc}
\end{subfigure}
\caption{Simulated environments. Panels (a)–(c): (a) Thin Obstacle; (b) Thin Obstacle with Rugged Terrain; (c) Dynamic Obstacle. Obstacle layouts are randomized at reset; only (c) updates obstacle positions during an episode.}
\label{fig:env_repre}
\end{figure}

Fig. \ref{fig:env_repre} shows representative examples. Unless noted otherwise, obstacle layouts are randomized at episode reset; only \emph{Dynamic Obstacle} updates obstacle positions during an episode.

\subsection{RL MDP Details}

\paragraph{Observation Space.}
The observation at time $t$ comprises a 93-dimensional proprioceptive vector $s^{\text{prop}}_t$ (IMU readings, local joint rotations, and the actions issued over the previous three time steps) and a stack of the four most recent dense depth images $\big[I^{\text{depth}}_{t-3},\, I^{\text{depth}}_{t-2},\, I^{\text{depth}}_{t-1},\, I^{\text{depth}}_{t}\big]$, each with resolution $64\times64$.

\paragraph{Action Space.}
The action is a 12-dimensional vector that specifies desired joint position targets for the 12 actuated joints. A proportional–derivative (PD) controller \citep{tan2009computation} converts these targets into joint torques on the agent.

\paragraph{Reward Function.}
A unified reward is used that balances forward progress, energy efficiency, and survival:
{\begin{equation}
R_t = \alpha_{\text{fwd}}\,R^{\text{fwd}}_t
     + \alpha_{\text{energy}}\,R^{\text{energy}}_t
     + \alpha_{\text{alive}}\,R^{\text{alive}}_t,
\end{equation}
where $\alpha_{\text{fwd}}{=}1$, $\alpha_{\text{energy}}{=}0.005$, $\alpha_{\text{alive}}{=}0.1$.}

The forward term encourages task-aligned motion,
\begin{equation}
R^{\text{fwd}}_t = \langle v_t,\ e_x\rangle
\end{equation}
where $v_t$ is the base linear velocity, $e_x$ is the unit vector along the $x$-axis.
The energy term penalizes excessive actuation,
\begin{equation}
R^{\text{energy}}_t = -\,\|\tau_t\|_2^2,
\end{equation}
where $\tau_t$ are realized joint torques.
The alive term rewards safe operation,
\begin{equation}
R^{\text{alive}}_t = 1\quad\text{until termination (falls or unrecoverable collisions)}.
\end{equation}

\subsection{Model Architecture Details}
As outlined in Fig. \ref{fig:overall_archi}, the network uses lightweight encoders, an SSM-based fusion backbone, and compact heads for control. Table~\ref{tab:arch} summarizes the architectural settings.

\begin{table}[t]
\centering
\small
\setlength{\tabcolsep}{6pt}
\caption{Architecture settings (compact).}
\label{tab:arch}
\begin{tabularx}{\linewidth}{
  @{\hspace{4pt}}                 
  >{\hspace{2pt}}l                
  >{\hspace{2pt}\raggedright\arraybackslash}X 
  @{\hspace{4pt}}                 
}
\toprule
\textbf{Component} & \textbf{Setting} \\
\midrule
Token width $d$ & 128 \\
Proprio encoder & 2-layer MLP (256, 256), ReLU \\
Visual token projection & CNN patchify $\rightarrow$ linear to width $d$ (=128) \\
Mamba backbone & $L_m{=}2$ stacked SSM layers, residual + LayerNorm \\
Projection head & 2-layer MLP (256, 256), ReLU \\
\bottomrule
\end{tabularx}
\end{table}

\subsection{Training Schema}
The model is trained with PPO using on-policy rollouts of length $T$, minibatch updates over several epochs, advantage normalization, and gradient clipping. Terrain and appearance randomization are applied at episode resets, and an obstacle-density curriculum gradually increases task difficulty. The model is evaluated periodically, repeated for 10 times, and the mean and standard deviation are reported over these runs. Table~\ref{tab:rl} lists PPO hyperparameters shared across methods. The overall LocoMamba training process, incorporating PPO, domain randomization, and curriculum learning, is illustrated in Algorithm~\ref{alg:train}.

\begin{table}[t]
\centering
\caption{PPO training hyperparameters.}
\label{tab:rl}
\begin{tabular*}{\linewidth}{
  @{\hspace{4pt}}              
  >{\hspace{2pt}}l             
  @{\extracolsep{\fill}}       
  >{\hspace{2pt}}l             
  @{\hspace{4pt}}              
}
\toprule
\textbf{Hyperparameter} & \textbf{Value} \\
\midrule
Episode horizon (steps) & 999 \\
Samples per iteration & 16{,}384 \\
Minibatch size & 1{,}024 \\
Optimization epochs per update & 3 \\
Discount factor $\gamma$ & 0.99 \\
GAE parameter $\lambda$ & 0.95 \\
PPO clip parameter $\epsilon$ & 0.2 \\
Entropy coefficient & 0.005 \\
Policy learning rate & $2\times 10^{-4}$ \\
Value learning rate & $2\times 10^{-4}$ \\
Optimizer & Adam \\
Nonlinearity & ReLU \\
\bottomrule
\end{tabular*}
\end{table}

\paragraph{Domain Randomization.}
To improve robustness, all methods use the same physics randomization \citep{mehta2020active} at episode reset, with parameters sampled uniformly from the ranges in Table~\ref{tab:dranges}. Unless otherwise noted, the randomized values remain fixed for the entire episode. 

\begin{table}[t]
\centering
\small
\setlength{\tabcolsep}{6pt}
\caption{Domain randomization ranges.}
\label{tab:dranges}
\begin{tabular*}{\linewidth}{
  @{\hspace{4pt}}
  @{\extracolsep{\fill}}       
  >{\hspace{2pt}}l
  >{\hspace{2pt}}l
  >{\hspace{2pt}}l
  @{\hspace{4pt}}
}
\toprule
\textbf{Parameter} & \textbf{Range} & \textbf{Units} \\
\midrule
$K_P$            & $[40,\ 90]$                 & - \\
$K_D$            & $[0.4,\ 0.8]$               & - \\
Link inertia     & $[0.5,\ 1.5]\times$ default & - \\
Lateral friction & $[0.5,\ 1.25]$              & N·s/m \\
Body mass        & $[0.8,\ 1.2]\times$ default & Kg \\
Motor friction   & $[0.0,\ 0.05]$              & N·m·s/rad \\
Motor strength   & $[0.8,\ 1.2]\times$ default & N·m \\
Sensor latency   & $[0,\ 0.04]$                & seconds \\
\bottomrule
\end{tabular*}
\end{table}

\paragraph{Visual Perturbations.}
In addition to physics randomization, the lightweight depth noise is also injected to simulate saturated returns and minor sensor artifacts. $K\sim\mathcal{U}\{3,\dots,30\}$ pixel locations are sampled per depth frame and their values are set to the sensor’s maximum range (salt-like saturation). This perturbation is applied consistently across all methods.

\paragraph{Obstacle-Density Curriculum.}
For obstacle-based scenarios, density starts from an easier setting and linearly ramps to the target distribution over training iterations \citep{wang2021survey}. Curriculum scheduling is identical across methods. {In our experiments, the obstacle density was gradually increased in a linear schedule from a sparse to a fully dense configuration, and the exact progression was determined empirically to maintain training stability. The purpose of this design is to introduce environmental complexity smoothly rather than abruptly, ensuring that the agent can progressively adapt to denser obstacle distributions.}

\paragraph{End-to-End Optimization.}
Gradients from the PPO objective flow through the projection head, the Mamba fusion backbone, and both encoders. The advantage normalization and gradient clipping are used and updated for several epochs per iteration with shuffled minibatches.

\begin{algorithm}[t]
\caption{LocoMamba training with PPO, domain randomization, and curriculum}
\label{alg:train}
\begin{algorithmic}[1]
\State \textbf{Input:} policy $\pi_\theta$, value $V_\phi$, fusion $f_{\text{fuse}}$, horizon $T$, samples/iter $N_{\text{iter}}$, PPO/GAE hyperparameters
\For{iteration $=1,2,\dots$}
  \State Sample obstacle density from curriculum; set scenario-specific textures/appearance
  \While{collected samples $< N_{\text{iter}}$}
    \State Sample physics params from Table~\ref{tab:dranges}; reset environment
    \For{$t=1$ to $T$}
      \State Observe $o_t=\{s^{\text{prop}}_t,\ \text{depth frames}\}$
      \State Inject depth perturbation: set $K\!\sim\!\mathcal{U}\{3,\dots,30\}$ random pixels $\rightarrow$ max range
      \State Compute fused feature $h_t=f_{\text{fuse}}(o_t)$ \Comment{Encoders + Mamba SSM}
      \State Sample action $\tilde{a}_t\!\sim\!\mathcal{N}(\mu_\theta(h_t),\text{diag}(\sigma_\theta^2(h_t)))$, set $a_t=\tanh(\tilde{a}_t)$
      \State Step env with $a_t$; receive $r_t$, $o_{t+1}$, done
      \State Store $(h_t,a_t,r_t,\log\pi_\theta(a_t|h_t), V_\phi(h_t),\text{done})$
      \If{done} \textbf{break} \EndIf
    \EndFor
  \EndWhile
  \State Compute advantages with GAE:
    $\delta_t=r_t+\gamma V_\phi(h_{t+1})-V_\phi(h_t)$,
    $A_t=\sum_{l\ge 0}(\gamma\lambda)^l\delta_{t+l}$
  \State Normalize advantages; compute returns $\hat{R}_t$
  \For{epoch $=1$ to $E$}
    \For{minibatch $\mathcal{B}$}
      \State $\rho_t=\frac{\pi_\theta(a_t|h_t)}{\pi_{\theta_{\text{old}}}(a_t|h_t)}$ for $t\in\mathcal{B}$
      \State $\mathcal{L}_{\text{clip}}=\mathbb{E}_{\mathcal{B}}\big[\min(\rho_t A_t,\ \text{clip}(\rho_t,1-\epsilon,1+\epsilon)A_t)\big]$
      \State $\mathcal{L}_V=\mathbb{E}_{\mathcal{B}}[(V_\phi(h_t)-\hat{R}_t)^2]$, \ \ $\mathcal{H}=\mathbb{E}_{\mathcal{B}}[\mathcal{H}(\pi_\theta(\cdot|h_t))]$
      \State Update $\theta,\phi$ to minimize $-\mathcal{L}_{\text{clip}} + \beta_V \mathcal{L}_V - \beta_H \mathcal{H}$ (with grad clipping)
    \EndFor
  \EndFor
  \State Advance curriculum schedule
\EndFor
\end{algorithmic}
\end{algorithm}

\section{Experimental Evaluation}
\label{sec:fifth}

\subsection{Evaluation Setup}

\paragraph{Research Questions}
Our evaluation is organized around the following research questions (RQs):
\begin{itemize}
  \item \textbf{RQ1 (SOTA performance).} Does \emph{LocoMamba}, which fuses proprioception and depth with Mamba SSM backbone, achieve better locomotion performance than SOTA, as measured by return, collision times, and distance moved?
  \item \textbf{RQ2 (Cross-modal effectiveness).} Does encoding both proprioception and depth yield better locomotion performance than using either modality alone?
  \item \textbf{RQ3 (Mamba SSM efficiency).} Under a fixed compute budget, does the Mamba-based fusion backbone reduce memory use and latency and reach target performance with fewer updates or less wall-clock time than attention-based or hierarchical fusion? 
  \item \textbf{RQ4 (Training robustness).} Does the proposed end-to-end PPO training scheme improve learning stability and performance consistency?

\end{itemize}

\paragraph{Evaluation Metrics}
The policies are evaluated using (i) \textit{mean episode return} and two domain-specific metrics: (ii) \textit{progress distance}, defined as the displacement in meters along the task-aligned target direction, and (iii) \textit{collision times}, computed over the entire locomotion ending either when 3 evaluation episodes complete or the robot falls. Collision checks are performed at every control step, and the collision metric is reported only for obstacles-bearing scenarios and for episodes that encounter at least one obstacle.

\paragraph{Baselines}
To assess the effectiveness of the proposed LocoMamba, comparisons are conducted with the following methods:

\begin{itemize}
  \item \textbf{Proprio-only.} Uses only the 93-D proprioceptive vector; no exteroception.
  \item \textbf{Proprio-Vision-Only.} Encodes proprioception and depth with the same encoders as ours; projects depth features to match the proprio feature width and concatenates the two vectors, which are fed directly to the policy/value heads (no sequence fusion).
  \item \textbf{Transformer Proprio-Vision.} Self-attention over visual tokens plus a proprio token, representing attention-based cross-modal fusion \citep{yang2021learning}. 
  \item \textbf{Transformer Vision-only.} Following Transformer Proprio-Vision, a transformer-based model using depth-only input is also evaluated.
\end{itemize}

All agents are trained using the same PPO schema, curriculum, and domain randomization, and are evaluated under identical settings. Where applicable, the same proprioceptive and depth encoders, as well as the same token width, are reused to minimize confounding factors related to representation size.

\subsection{Performance vs.\ Baselines on Simulation Scenarios}
\label{sec:mainresults}
\begin{figure}[t]
\centering
\includegraphics[width=\linewidth]{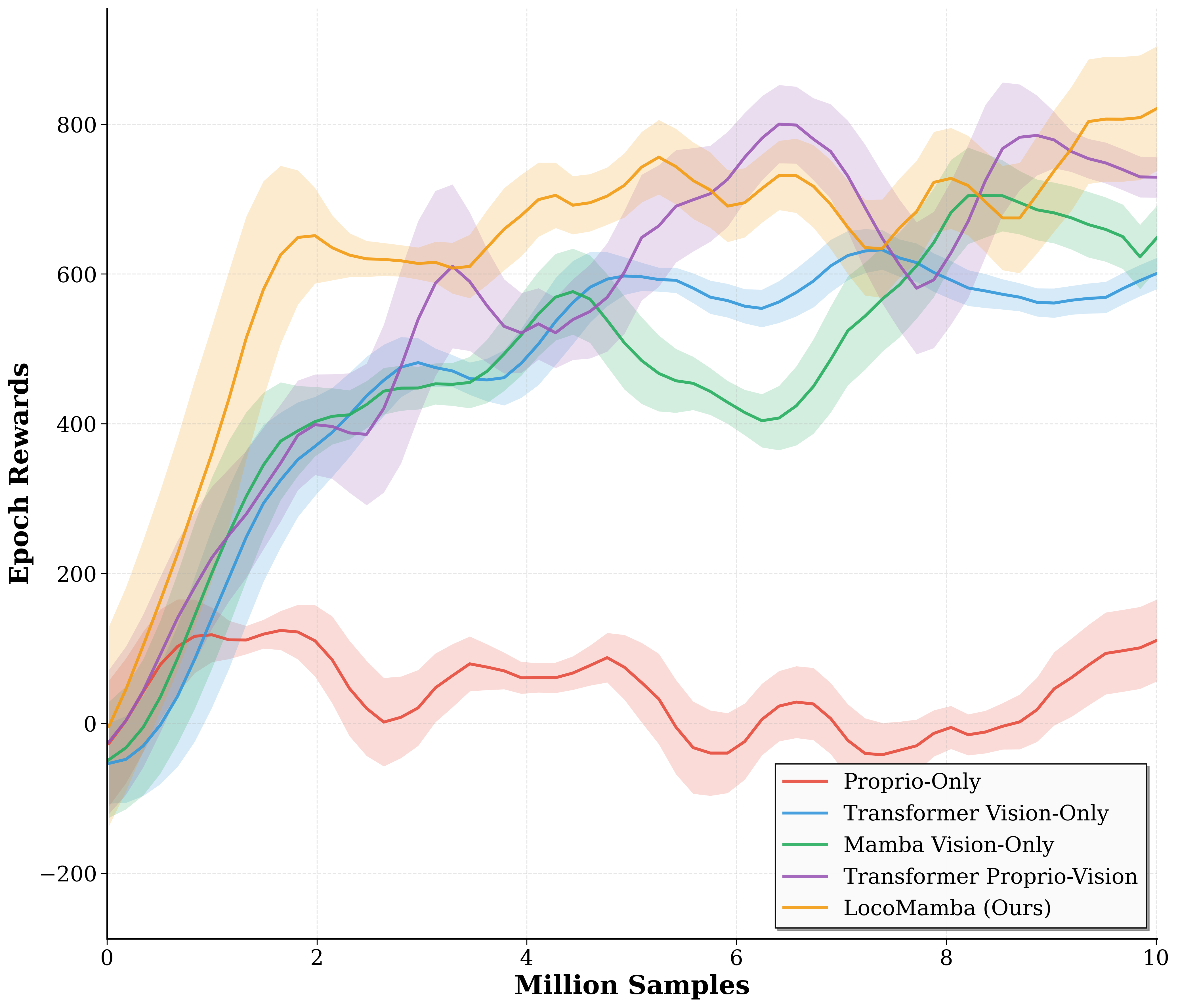}
\caption{{Training learning curves on \emph{Thin Obstacle}. Solid lines denote the mean episode return across seeds; shaded regions indicate $\pm$ one standard deviation.}}
\label{fig:learn_curves}
\end{figure}

\emph{LocoMamba} and its variant, \emph{Mamba Vision-Only}, which uses only vision as input, are evaluated against established baselines in the \emph{Thin Obstacle} environment. Fig. \ref{fig:learn_curves} shows training curves with mean and one standard deviation across seeds. Table~\ref{tab:performance_comparison} reports means and standard deviations over seeds for episode return, collisions per episode, and progress distance.

From Fig.~\ref{fig:learn_curves}, \emph{LocoMamba} exhibits faster convergence (steeper early learning slope), higher asymptotic return, and lower variance than all baselines. These trends substantiate its SOTA performance, answering \textbf{RQ1}, and indicate stable optimization under the proposed PPO training protocol, addressing \textbf{RQ4}.

\begin{table*}[t]
\centering
\begin{minipage}{0.92\linewidth} 
  \centering
  \caption{Performance on the \emph{Thin Obstacle} environment. Mean $\pm$ std over 10 runs.}
  \label{tab:performance_comparison}
  \small
  \begin{tabular*}{\linewidth}{  
    @{\hspace{4pt}}              
    l
    @{\extracolsep{\fill}}       
    c c c
    @{\hspace{4pt}}              
  }
  \toprule
  \textbf{Model Architecture} & \textbf{Return} &
  \textbf{Collision Times} & \textbf{Distance Moved} \\
  \midrule
  Proprio-Only               & 145.64 $\pm$ 89.55  & 487.80 $\pm$ 114.13 & 6.13 $\pm$ 2.49 \\
  Transformer Vision-Only    & 187.45 $\pm$ 93.88  & --                  & 6.75 $\pm$ 2.90 \\
  Mamba Vision-Only          & 28.16  $\pm$ 34.47  & --                  & 2.92 $\pm$ 1.01 \\
  Transformer Proprio-Vision & 511.96 $\pm$ 247.30 & 141.83 $\pm$ 158.47 & 24.85 $\pm$ 7.34 \\
  LocoMamba (Ours)           & \textbf{762.34 $\pm$ 156.53} & \textbf{72.53 $\pm$ 79.47} & \textbf{32.41 $\pm$ 5.23} \\
  \bottomrule
  \end{tabular*}
\end{minipage}
\end{table*}

From Table~\ref{tab:performance_comparison}, \emph{Mamba Proprio-Vision} attains the best overall performance, with higher return ($762.34\pm156.53$), fewer collisions ($72.53\pm79.47$), and longer progress distance ($32.41\pm5.23$) than all baselines. Relative to \emph{Transformer Proprio-Vision}, it improves return by $48.9\%$, reduces collisions by $48.9\%$, and increases distance by $30.4\%$. Compared with \emph{Proprio-Only}, it yields a $423.4\%$ gain in return, an $85.1\%$ reduction in collisions, and a $428.7\%$ increase in distance. The two vision-only variants move little on average, so collision counts are not informative and are omitted. These results indicate that cross-modal fusion is essential for obstacle negotiation and that the selective state-space backbone confers a clear advantage under identical training and evaluation protocols. The findings answer \textbf{RQ1} affirmatively and support the effectiveness of \emph{LocoMamba}. Moreover, under the proposed end-to-end PPO training protocol with a compact state-centric reward, the agent exhibits stable learning and balanced trade-offs among task-aligned progress, energy efficiency, and safety, thereby answering \textbf{RQ4}.

\subsection{Ablation Studies}
\label{sec:ablations}

\paragraph{Ablation on Modalities.}
From Table~\ref{tab:performance_comparison}, \textit{Transformer Vision-State} improves over \textit{Proprio-Only} by about 251.5\% in return, reduces collisions by about 70.9\%, and increases progress distance by about 305.4\%. It also outperforms \textit{Transformer Vision-Only} with about 173.1\% higher return and about 268.1\% greater distance. \textit{LocoMamba} further widens these margins: relative to \textit{Proprio-Only} it achieves about 423.4\% higher return, about 85.1\% fewer collisions, and about 428.7\% longer distance; compared with \textit{Mamba Vision-Only} it delivers roughly 27.1-fold the return and 11.1-fold the distance. Collision counts for vision-only variants are omitted because these policies move too little for the metric to be informative. These results demonstrate the effectiveness of combining proprioception and depth and answer \textbf{RQ2}.

\paragraph{Ablation on the Cross-Modal Fusion Backbone.}
\label{sec:ablations-backbone}
The Mamba-based cross-modal backbone is now evaluated for both effectiveness and efficiency through controlled ablations against baselines.\\
\noindent\textbf{Effectiveness of the Mamba cross-modal backbone.}
The fusion backbones is evaluated under matched modality settings. In the proprioception–vision setting, \textit{LocoMamba} outperforms \textit{Transformer Proprio-Vision} on all metrics in Table~\ref{tab:performance_comparison}. The return rises from $511.96\pm247.30$ to $762.34\pm156.53$ (a $48.9\%$ increase), the number of collisions drops from $141.83\pm158.47$ to $72.53\pm79.47$ (a $48.9\%$ reduction), and the progress distance increases from $24.85\pm7.34$ to $32.41\pm5.23$ (a $30.4\%$ increase). In the vision-only setting, both methods perform poorly relative to cross-modal policies. \textit{Mamba Vision-Only} remains substantially below \textit{Transformer Vision-Only} in return and distance, underscoring the importance of proprioception for egocentric perception and control. {The lower performance of the Mamba Vision-Only model arises from its higher temporal sensitivity: without proprioceptive input, its recurrent state gradually drifts under partial observability. In contrast, the Transformer’s self-attention captures limited historical context within each visual window. This highlights that Mamba’s selective recurrence benefits most from proprioceptive grounding rather than indicating a structural weakness.} These comparisons indicate that the Mamba SSM backbone is most beneficial when fusing modalities within a single policy. 

\noindent\textbf{Efficiency of the Mamba cross-modal backbone.}
Table~\ref{tab:learning_analysis} summarizes learning-curve statistics under identical training schedules. In the proprioception–vision setting, the Mamba backbone achieves a higher final reward (737.4 compared with 714.1), a steeper early-learning slope (6.41 per epoch compared with 3.66, a $75\%$ increase), higher learning efficiency (1.44 compared with 1.28), and a larger area under the learning curve per epoch (601.7 compared with 529.8). In the vision-only setting, the Mamba backbone attains a higher final reward (676.0 compared with 577.8) and a steeper early slope (4.22 compared with 3.69), with a comparable area under the curve. Taken together, these results support \textbf{RQ3}: the Mamba-based backbone improves optimization dynamics under the same training budget and confers efficiency benefits that translate into faster learning and stronger final performance when cross-modal fusion is used. In addition, the learning curves in Fig.~\ref{fig:learn_curves} show faster convergence and lower variance for the Mamba-based models, indicating more stable optimization. {As shown in Table \ref{tab:efficiency}, the proposed LocoMamba model achieves the highest throughput (104.7 samples/s) while maintaining a lower allocated memory footprint (207 MB) compared to its Transformer-based counterpart (226 MB). Moreover, despite having slightly fewer parameters (300K vs. 353K), LocoMamba exhibits comparable FLOPs per step, demonstrating that its superior runtime efficiency results from architectural optimization rather than reduced computational load.} This further supports the efficiency advantage of the Mamba SSM backbone (\textbf{RQ3}) and the robustness of the proposed PPO training protocol (\textbf{RQ4}).

\begin{table*}[h]
{
\centering
\begin{minipage}{\linewidth} 
  \centering
  \begin{threeparttable}
    \caption{Learning Efficiency Analysis: Performance and Speed}
    \label{tab:learning_analysis}
    \setlength{\tabcolsep}{1pt}
    \small
    \begin{tabular*}{\linewidth}{   
      @{\hspace{1pt}}               
      l
      @{\extracolsep{\fill}}        
      c c c c
      @{\hspace{1pt}}               
    }
    \toprule
    \textbf{Model} & \textbf{Return} &
    \textbf{\shortstack{Early\\Learning Slope}} &
    \textbf{\shortstack{Learning\\Efficiency}} &
    \textbf{\shortstack{AUC\\per Epoch}} \\
    \midrule
    Proprio-Only               & 27.2  & 1.69 & 0.27 & 36.5  \\
    Transformer Vision-Only    & 577.8 & 3.69 & 1.07 & 448.3 \\
    Mamba Vision-Only          & 676.0 & 4.22 & 1.15 & 443.8 \\
    Transformer Proprio-Vision & 714.1 & 3.66 & 1.28 & 529.8 \\
    LocoMamba (Ours)           & \textbf{737.4} & \textbf{6.41} & \textbf{1.44} & \textbf{601.7} \\
    \bottomrule
    \end{tabular*}

    \begin{tablenotes}[para,flushleft]
    \footnotesize
    \item Final Reward: average reward over the last 120 epochs (2M samples).
    Early Learning Slope: reward gain per epoch (first 120 epochs).
    Learning Efficiency: overall reward gain per epoch.
    AUC: area under the curve per epoch. Best values are in bold.
    \end{tablenotes}
  \end{threeparttable}
\end{minipage}
}
\end{table*}

\begin{table*}[h]
{
\centering
\begin{minipage}{\linewidth} 
  \centering
  \begin{threeparttable}
    \caption{Computational and Memory Efficiency Comparison between Mamba-based and Transformer-based Models}
    \label{tab:efficiency}
    \setlength{\tabcolsep}{1pt} 
    {\scriptsize
    \begin{tabular*}{\linewidth}{
      @{\hspace{1pt}}
      l
      @{\extracolsep{\fill}}
      c c c c c c c
      @{\hspace{1pt}}
    }
    \toprule
    \textbf{Model} &
    \textbf{\shortstack{Reserved\\Memory\\(MB)}} &
    \textbf{\shortstack{Allocated\\Memory\\(MB)}} &
    \textbf{Samples/s} &
    \textbf{\shortstack{FLOPs/step\\(Policy)}} &
    \textbf{\shortstack{FLOPs/step\\(Value)}} &
    \textbf{\shortstack{Params\\(Policy)}} &
    \textbf{\shortstack{Params\\(Value)}} \\
    \midrule
    {LocoMamba (Ours)} & {3160} & {207.2} & {104.68} & {4113664} & {4112384} & {300070} & {298785} \\
    Transformer Proprio-Vision    & 3156 & 226.4 & 98.87  & 5012992 & 5011712 & 353190 & 351905 \\
    \midrule
    Mamba Vision-Only       & 3098 & 189.6 & 100.35 & 4198912 & 4197632 & 193254 & 191969 \\
    Transformer Vision-Only  & 3022 & 199.2 & 102.84 & 4826624 & 4825344 & 232806 & 231521 \\
    \bottomrule
    \end{tabular*}
    }
  \end{threeparttable}
  \vspace{2pt}
\end{minipage}
}
\end{table*}

{
To further quantify the trade-off between spatial resolution and computational efficiency, 
we conducted an ablation study by varying the number of visual tokens 
(\(32\), \(64\), and \(256\)) generated from the CNN patchification process. 
The corresponding performance and efficiency results are reported in 
Tables~\ref{tab:perf} and~\ref{tab:effi}, respectively. 
As expected, increasing the token count provides the model with finer spatial granularity, 
which improves perception accuracy in trained environments. 
However, it also leads to higher computational overhead due to the quadratic growth of 
feature representations and the additional intermediate states required in the Mamba layers. 
Among the tested settings, the \(256\)-token configuration yields the highest return and 
distance on the trained (thin obstacle) terrain but exhibits greater memory usage and 
degraded generalization on unseen rugged terrains. 
In contrast, the \(64\)-token configuration achieves a favorable balance, offering 
competitive performance across all environments while maintaining moderate GPU memory 
consumption and FLOPs. 
These results indicate that an intermediate token size provides the best trade-off 
between spatial fidelity, computational efficiency, and cross-domain robustness. 
}

\begin{table*}[t]
\centering
\begin{minipage}{0.92\linewidth}
  \centering
  \begin{threeparttable}
    \caption{{Performance comparison across different visual token configurations (32, 64, 256).}}
    \label{tab:perf}
    \setlength{\tabcolsep}{5pt}
    \small
    {
    \begin{tabular*}{\linewidth}{
      @{\hspace{4pt}}
      l
      @{\extracolsep{\fill}}
      c c c
      @{\hspace{4pt}}
    }
    \toprule
    \textbf{Token Size} & \textbf{Reward} & \textbf{Collisions} & \textbf{Distance} \\
    \midrule
    \multicolumn{4}{c}{\textit{Thin Obstacle Environment}} \\
    \midrule
    32  & 740.55 $\pm$ 193.45 & \textbf{57.88 $\pm$ 39.77} & 32.93 $\pm$ 7.39 \\
    64 (Ours) & 762.34 $\pm$ 156.53 & 72.53 $\pm$ 79.47 & 32.41 $\pm$ 5.23 \\
    256 & \textbf{797.46 $\pm$ 111.29} & 106.95 $\pm$ 92.71 & \textbf{34.74 $\pm$ 5.02} \\
    \midrule
    \multicolumn{4}{c}{\textit{Dynamic Obstacle Environment}} \\
    \midrule
    32  & 697.33 $\pm$ 587.19 & \textbf{12.31 $\pm$ 89.88} & 29.39 $\pm$ 8.00 \\
    64 (Ours) & \textbf{749.09 $\pm$ 269.19} & 38.47 $\pm$ 43.96 & \textbf{32.51 $\pm$ 5.84} \\
    256 & 682.21 $\pm$ 569.84 & 260.51 $\pm$ 203.64 & 29.08 $\pm$ 9.81 \\
    \midrule
    \multicolumn{4}{c}{\textit{Rugged Unseen Environment}} \\
    \midrule
    32  & 563.39 $\pm$ 166.65 & 1842.38 $\pm$ 357.57 & 23.97 $\pm$ 5.78 \\
    64 (Ours) & \textbf{583.76 $\pm$ 154.57} & \textbf{470.27 $\pm$ 108.15} & \textbf{24.04 $\pm$ 6.18} \\
    256 & 562.61 $\pm$ 185.64 & 4166.64 $\pm$ 708.69 & 23.59 $\pm$ 6.70 \\
    \bottomrule
    \end{tabular*}
    }
    
  \end{threeparttable}
\end{minipage}
\end{table*}

\begin{table*}[t]
\centering
\begin{minipage}{0.92\linewidth}
  \centering
  \begin{threeparttable}
    \caption{{Efficiency comparison across different visual token configurations (32, 64, 256).}}
    \label{tab:effi}
    \setlength{\tabcolsep}{5pt}
    \small
    {
    \begin{tabular*}{\linewidth}{
      @{\hspace{4pt}}
      c
      @{\extracolsep{\fill}}
      c c c
      @{\hspace{4pt}}
    }
    \toprule
    \textbf{Token Size} & \textbf{\shortstack{Memory Usage\\(Reserved / Allocated)}} & 
    \textbf{\shortstack{FLOPs\\(Policy / Value)}} & 
    \textbf{\shortstack{Params\\(Policy / Value)}} \\
    \midrule
    32  & 2980 MB / 200 MB & 3{,}901{,}952 / 3{,}903{,}232 & 263{,}137 / 264{,}422 \\
    64  & 3160 MB / 207.2 MB & 4{,}112{,}384 / 4{,}113{,}664 & 298{,}785 / 300{,}070 \\
    256 & 3546 MB / 359.6 MB & 6{,}928{,}640 / 6{,}929{,}920 & 604{,}065 / 605{,}350 \\
    \bottomrule
    \end{tabular*}
    }
   
  \end{threeparttable}
\end{minipage}
\end{table*}

\subsection{Learning Stability Analysis}
\label{sec:stability}
\begin{figure}[t]
\centering
\includegraphics[width=\linewidth]{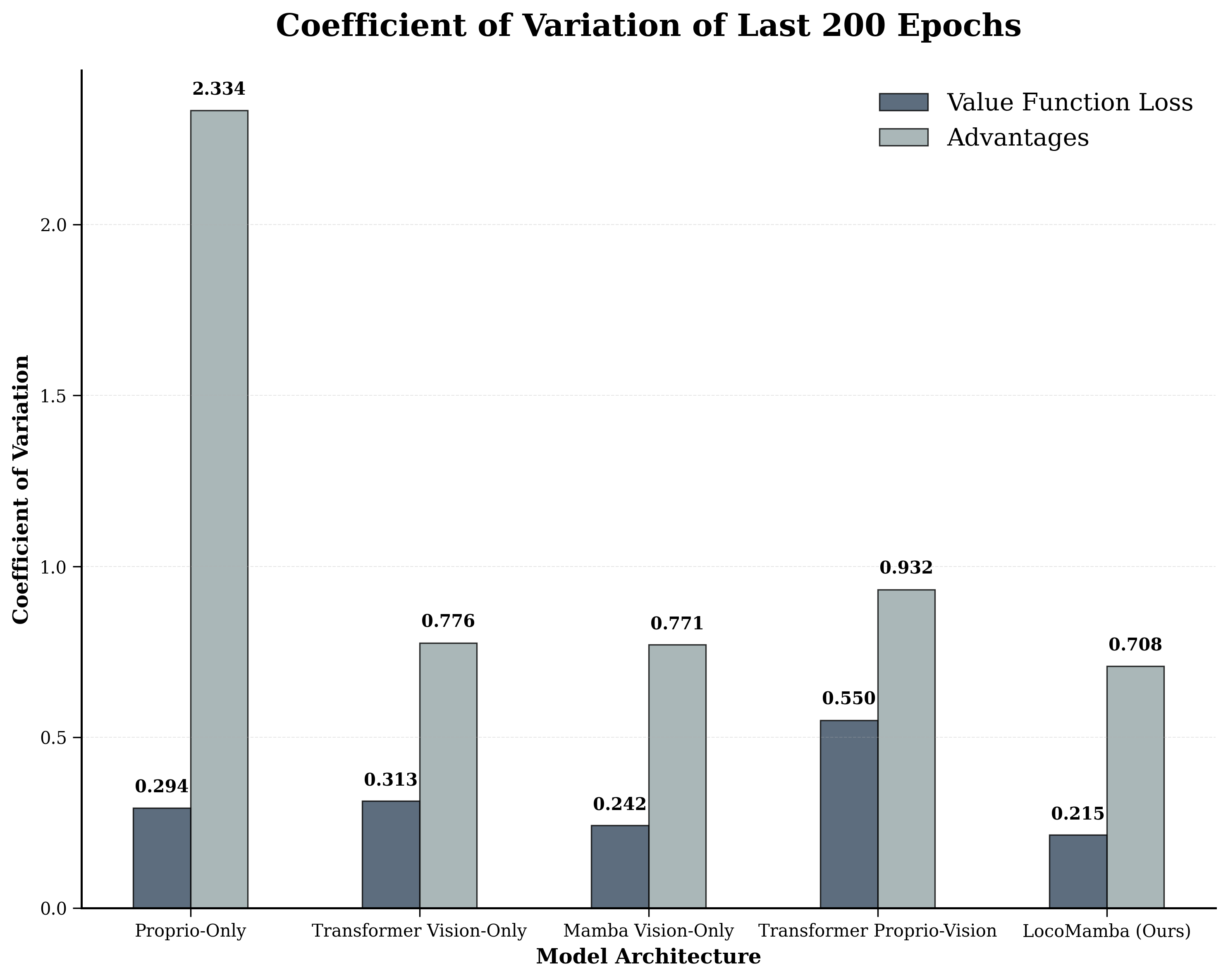} 
\caption{Coefficient of variation over the last 200 epochs. Bars show CoV for the value-function loss (dark) and advantages (light). Lower values indicate more stable optimization.}
\label{fig:cov_last200}
\end{figure}

Fig. \ref{fig:cov_last200} reports the coefficient of variation (CoV = std/mean) over the last 200 training epochs for the value-function loss and the estimated advantages. Lower is better.

\emph{LocoMamba} (Proprio–Vision) attains the lowest CoV on both metrics (value-loss CoV $0.215$, advantage CoV $0.708$), indicating the most stable optimization. Relative to \emph{Transformer Proprio–Vision}, the value-loss CoV decreases from $0.550$ to $0.215$ (a $\sim\!61\%$ reduction), and the advantage CoV decreases from $0.932$ to $0.708$ (a $\sim\!24\%$ reduction). Vision-only policies exhibit moderate stability but limited displacement, while \emph{Proprio-Only} shows highly volatile advantages (CoV $2.334$), consistent with partial observability.

These results corroborate the efficiency and stability benefits of the Mamba SSM backbone (\textbf{RQ3}) and demonstrate that the proposed PPO training protocol yields stable learning with balanced updates (\textbf{RQ4}).

\subsection{Generalization to Unseen Conditions}
\label{sec:generalization}
The policies are trained in the \emph{Thin Obstacle} environment and evaluate them zero-shot on two previously unseen scenarios: a rugged terrain and a dynamic-obstacle terrain with randomly moving hazards. Tables~\ref{tab:model_comparison} and \ref{tab:performance_comparison_moving} report means and standard deviations over 10 runs. For vision-only policies, collision counts are omitted because the agents move too little for this metric to be informative.

\begin{table*}[t]
\centering
\begin{minipage}{0.92\linewidth}
  \centering
  \caption{Zero-shot performance on a rugged unseen terrain (trained on \emph{Thin Obstacle}). Mean $\pm$ std over 10 runs.}
  \label{tab:model_comparison}
  \small
  \begin{tabular*}{\linewidth}{
    @{\hspace{4pt}} l @{\extracolsep{\fill}} c c c @{\hspace{4pt}}
  }
  \toprule
  \textbf{Model Architecture} & \textbf{Return} & \textbf{Collision Times} & \textbf{Distance Moved} \\
  \midrule
  Proprio-Only               & 106.45 $\pm$ 74.94  & 779.40 $\pm$ 119.85 & 4.94 $\pm$ 1.97 \\
  Transformer Vision-Only    & 151.75 $\pm$ 82.28  & --                  & 5.75 $\pm$ 2.68 \\
  Mamba Vision-Only          & 28.53  $\pm$ 29.25  & --                  & 2.79 $\pm$ 0.96 \\
  Transformer Proprio-Vision & 257.38 $\pm$ 431.52 & 593.73 $\pm$ 135.62 & 16.45 $\pm$ 8.25 \\
  LocoMamba (Ours)           & \textbf{583.76 $\pm$ 154.57} & \textbf{470.27 $\pm$ 108.15} & \textbf{24.04 $\pm$ 6.18} \\
  \bottomrule
  \end{tabular*}
\end{minipage}
\end{table*}

\begin{table*}[t]
\centering
\begin{minipage}{0.92\linewidth}
  \centering
  \caption{Zero-shot performance on a dynamic-obstacle terrain (trained on \emph{Thin Obstacle}). Mean $\pm$ std over 10 runs.}
  \label{tab:performance_comparison_moving}
  \small
  \begin{tabular*}{\linewidth}{
    @{\hspace{4pt}} l @{\extracolsep{\fill}} c c c @{\hspace{4pt}}
  }
  \toprule
  \textbf{Model Architecture} & \textbf{Return} & \textbf{Collision Times} & \textbf{Distance Moved} \\
  \midrule
  Proprio-Only               & 119.20 $\pm$ 26.33  & 423.03 $\pm$ 187.08 & 5.01 $\pm$ 0.88 \\
  Transformer Vision-Only    & 196.78 $\pm$ 44.64  & --                  & 7.49 $\pm$ 1.51 \\
  Mamba Vision-Only          & 19.32  $\pm$ 26.52  & --                  & 2.57 $\pm$ 0.79 \\
  Transformer Proprio-Vision & 685.49 $\pm$ 133.31 & 121.63 $\pm$ 83.15  & 28.36 $\pm$ 4.77 \\
  LocoMamba (Ours)           & \textbf{749.09 $\pm$ 269.19} & \textbf{38.47 $\pm$ 43.96} & \textbf{32.51 $\pm$ 5.84} \\
  \bottomrule
  \end{tabular*}
\end{minipage}
\end{table*}

\paragraph{Rugged Terrain.}
Cross-modal policies generalize substantially better than single-modality baselines. \emph{LocoMamba} achieves the highest return and progress with the fewest collisions. Relative to \emph{Proprio-Only}, return increases by approximately $448\%$, collisions decrease by $39.7\%$, and distance increases by $386.6\%$. Compared with \emph{Transformer Proprio-Vision}, \emph{LocoMamba} attains a $126.8\%$ higher return, $20.8\%$ fewer collisions, and a $46.1\%$ longer distance. Vision-only variants travel little, indicating limited generalization without proprioceptive anchoring. These results show that combining proprioception with depth is effective for out-of-distribution generalization and that the selective state-space backbone strengthens cross-modal fusion under distribution shift.

\paragraph{Dynamic obstacles.}
\emph{LocoMamba} maintains a clear advantage when dynamically moving. Relative to \emph{Transformer Proprio-Vision}, return increases by about $9\%$, collisions are reduced by roughly two thirds, and distance increases by about $15\%$. Relative to \emph{Proprio-Only}, return and distance increase by more than a factor of six while collisions fall by approximately $91\%$. Vision-only policies again show limited displacement. 

These findings indicate that the proposed PPO-trained cross-modal Mamba design preserves performance under temporal disturbances and moving hazards, demonstrating strong generalization to previously unseen challenging and dynamic environments. Taken together, the results address \textbf{RQ1} to \textbf{RQ4} by confirming overall performance gains, the effectiveness of combining proprioception and depth, the efficiency of the Mamba state-space backbone, and the robustness of the training protocol.

\section{Conclusion}
\label{sec:sixth}
This paper presented \emph{LocoMamba}, a vision-driven cross-modal reinforcement learning framework for quadrupedal locomotion. The method embeds proprioceptive state with a lightweight MLP and patchifies depth images with a compact CNN, then fuses the resulting tokens using stacked Mamba selective state-space layers. Policies and values are optimized end to end with PPO under terrain and appearance randomization and an obstacle-density curriculum, guided by a compact state-centric reward.

Across challenging simulated environments, \emph{LocoMamba} achieved higher returns, fewer collisions, and longer progress distances than strong baselines that include proprioception-only controllers, depth-only controllers, and Transformer-based fusion. Learning-curve analyses and efficiency metrics showed faster convergence and improved sample efficiency under the same compute budget, consistent with the near-linear scaling of the selective state-space backbone. Stability indicators, including variability of advantages and value loss, further confirmed reliable optimization.

Ablation studies established the importance of combining proprioception with depth and showed that the Mamba backbone is particularly effective for cross-modal fusion. The proposed PPO training protocol, which combines domain randomization, curriculum scheduling, and a compact reward, produced stable learning and balanced trade-offs among task-aligned progress, energy efficiency, and safety. Together, these results answer the research questions on overall performance, efficiency, modality design, and training robustness in the affirmative.

{Due to current budgetary and hardware-access constraints, the real-world experiments have not been conducted. When resources permit, \emph{LocoMamba} will be deployed on a quadruped platform and assess sim-to-real transfer, latency, and safety under field conditions.} {Future work will incorporate more realistic visual perturbations such as occlusion holes, depth bias, and motion blur, as well as temporal latency and misalignment, to further evaluate the robustness of the Mamba-based cross-modal fusion in real-world conditions.}

\printcredits

\section*{Declaration of competing interest}
The authors declare that they have no known competing financial interests or personal relationships that could have appeared to influence the work reported in this paper.

\section*{Acknowledgment}
This research was funded by the Graduate Innovation Fund of Jilin University under Grant 101832020CX130.

\bibliographystyle{model1-num-names}

\bibliography{cas-refs}

@article{fan2024review,
  title={A review of quadruped robots: Structure, control, and autonomous motion},
  author={Fan, Yanan and Pei, Zhongcai and Wang, Chen and Li, Meng and Tang, Zhiyong and Liu, Qinghua},
  journal={Advanced Intelligent Systems},
  volume={6},
  number={6},
  pages={2300783},
  year={2024},
  publisher={Wiley Online Library}
}

@article{carpentier2021recent,
  title={Recent progress in legged robots locomotion control},
  author={Carpentier, Justin and Wieber, Pierre-Brice},
  journal={Current Robotics Reports},
  volume={2},
  number={3},
  pages={231--238},
  year={2021},
  publisher={Springer}
}

@article{xin2021robust,
  title={Robust footstep planning and LQR control for dynamic quadrupedal locomotion},
  author={Xin, Guiyang and Xin, Songyan and Cebe, Oguzhan and Pollayil, Mathew Jose and Angelini, Franco and Garabini, Manolo and Vijayakumar, Sethu and Mistry, Michael},
  journal={IEEE Robotics and Automation Letters},
  volume={6},
  number={3},
  pages={4488--4495},
  year={2021},
  publisher={IEEE}
}

@article{zhang2022deepreinforcementlearningforreal,
  title={Deepreinforcementlearningforreal-world quadrupedal locomotion: a comprehensive review},
  author={Zhang, Hongyin and He, Li and Wang, Donglin},
  year={2022}
}

@inproceedings{xie2021dynamics,
  title={Dynamics randomization revisited: A case study for quadrupedal locomotion},
  author={Xie, Zhaoming and Da, Xingye and Van de Panne, Michiel and Babich, Buck and Garg, Animesh},
  booktitle={2021 IEEE International Conference on Robotics and Automation (ICRA)},
  pages={4955--4961},
  year={2021},
  organization={IEEE}
}

@article{lee2020learning,
  title={Learning quadrupedal locomotion over challenging terrain},
  author={Lee, Joonho and Hwangbo, Jemin and Wellhausen, Lorenz and Koltun, Vladlen and Hutter, Marco},
  journal={Science robotics},
  volume={5},
  number={47},
  pages={eabc5986},
  year={2020},
  publisher={American Association for the Advancement of Science}
}

@inproceedings{xie2022glide,
  title={Glide: Generalizable quadrupedal locomotion in diverse environments with a centroidal model},
  author={Xie, Zhaoming and Da, Xingye and Babich, Buck and Garg, Animesh and de Panne, Michiel van},
  booktitle={International workshop on the algorithmic foundations of robotics},
  pages={523--539},
  year={2022},
  organization={Springer}
}

@article{han2025multimodal,
  title={Multimodal fusion and vision-language models: A survey for robot vision},
  author={Han, Xiaofeng and Chen, Shunpeng and Fu, Zenghuang and Feng, Zhe and Fan, Lue and An, Dong and Wang, Changwei and Guo, Li and Meng, Weiliang and Zhang, Xiaopeng and others},
  journal={arXiv preprint arXiv:2504.02477},
  year={2025}
}

@inproceedings{li2024rapid,
  title={Rapid Learning of Natural Gaits for Quadrupedal Locomotion and Skill Reuse in Downstream Tasks},
  author={Li, Kaicen and Gao, Wei and Zhang, Shiwu},
  booktitle={2024 IEEE International Conference on Robotics and Biomimetics (ROBIO)},
  pages={2330--2336},
  year={2024},
  organization={IEEE}
}

@inproceedings{habu2018simple,
  title={A simple rule for quadrupedal gait transition proposed by a simulated muscle-driven quadruped model with two-level cpgs},
  author={Habu, Yasushi and Yamada, Yuuta and Fukui, Satoshi and Fukuoka, Yasuhiro},
  booktitle={2018 IEEE International Conference on robotics and biomimetics (ROBIO)},
  pages={2075--2081},
  year={2018},
  organization={IEEE}
}

@inproceedings{bledt2018cheetah,
  title={Mit cheetah 3: Design and control of a robust, dynamic quadruped robot},
  author={Bledt, Gerardo and Powell, Matthew J and Katz, Benjamin and Di Carlo, Jared and Wensing, Patrick M and Kim, Sangbae},
  booktitle={2018 IEEE/RSJ International Conference on Intelligent Robots and Systems (IROS)},
  pages={2245--2252},
  year={2018},
  organization={IEEE}
}

@article{liu2014planning,
  title={Planning and simulation of the rule-based trotting gait of a bionic quadruped robot},
  author={Liu, Ze Guo and Ding, Xiang Fang},
  journal={Advanced Materials Research},
  volume={971},
  pages={624--628},
  year={2014},
  publisher={Trans Tech Publ}
}

@article{miura1984dynamic,
  title={Dynamic walk of a biped},
  author={Miura, Hirofumi and Shimoyama, Isao},
  journal={The International Journal of Robotics Research},
  volume={3},
  number={2},
  pages={60--74},
  year={1984},
  publisher={Sage Publications Sage CA: Thousand Oaks, CA}
}

@inproceedings{yao2021hierarchy,
  author={Yao, Qingfeng and Wang, Jilong and Wang, Donglin and Yang, Shuyu and Zhang, Hongyin and Wang, Yinuo and Wu, Zhengqing},
  booktitle={2021 IEEE/RSJ International Conference on Intelligent Robots and Systems (IROS)}, 
  title={Hierarchical Terrain-Aware Control for Quadrupedal Locomotion by Combining Deep Reinforcement Learning and Optimal Control}, 
  year={2021},
  volume={},
  number={},
  pages={4546-4551}
}

@inproceedings{di2018dynamic,
  title={Dynamic locomotion in the mit cheetah 3 through convex model-predictive control},
  author={Di Carlo, Jared and Wensing, Patrick M and Katz, Benjamin and Bledt, Gerardo and Kim, Sangbae},
  booktitle={2018 IEEE/RSJ international conference on intelligent robots and systems (IROS)},
  pages={1--9},
  year={2018},
  organization={IEEE}
}

@inproceedings{ding2019real,
  title={Real-time model predictive control for versatile dynamic motions in quadrupedal robots},
  author={Ding, Yanran and Pandala, Abhishek and Park, Hae-Won},
  booktitle={2019 International Conference on Robotics and Automation (ICRA)},
  pages={8484--8490},
  year={2019},
  organization={IEEE}
}

@article{carius2019trajectory,
  title={Trajectory optimization for legged robots with slipping motions},
  author={Carius, Jan and Ranftl, Ren{\'e} and Koltun, Vladlen and Hutter, Marco},
  journal={IEEE Robotics and Automation Letters},
  volume={4},
  number={3},
  pages={3013--3020},
  year={2019},
  publisher={IEEE}
}

@inproceedings{grandia2019feedback,
  title={Feedback mpc for torque-controlled legged robots},
  author={Grandia, Ruben and Farshidian, Farbod and Ranftl, Ren{\'e} and Hutter, Marco},
  booktitle={2019 IEEE/RSJ International Conference on Intelligent Robots and Systems (IROS)},
  pages={4730--4737},
  year={2019},
  organization={IEEE}
}

@article{elobaid2025adaptive,
  title={Adaptive non-linear centroidal mpc with stability guarantees for robust locomotion of legged robots},
  author={Elobaid, Mohamed and Turrisi, Giulio and Rapetti, Lorenzo and Romualdi, Giulio and Dafarra, Stefano and Kawakami, Tomohiro and Chaki, Tomohiro and Yoshiike, Takahide and Semini, Claudio and Pucci, Daniele},
  journal={IEEE Robotics and Automation Letters},
  year={2025},
  publisher={IEEE}
}

@inproceedings{amatucci2024accelerating,
  title={Accelerating model predictive control for legged robots through distributed optimization},
  author={Amatucci, Lorenzo and Turrisi, Giulio and Bratta, Angelo and Barasuol, Victor and Semini, Claudio},
  booktitle={2024 IEEE/RSJ International Conference on Intelligent Robots and Systems (IROS)},
  pages={12734--12741},
  year={2024},
  organization={IEEE}
}

@article{hwangbo2019learning,
  title={Learning agile and dynamic motor skills for legged robots},
  author={Hwangbo, Jemin and Lee, Joonho and Dosovitskiy, Alexey and Bellicoso, Dario and Tsounis, Vassilios and Koltun, Vladlen and Hutter, Marco},
  journal={Science Robotics},
  volume={4},
  number={26},
  pages={eaau5872},
  year={2019},
  publisher={American Association for the Advancement of Science}
}

@article{tan2018sim,
  title={Sim-to-real: Learning agile locomotion for quadruped robots},
  author={Tan, Jie and Zhang, Tingnan and Coumans, Erwin and Iscen, Atil and Bai, Yunfei and Hafner, Danijar and Bohez, Steven and Vanhoucke, Vincent},
  journal={arXiv preprint arXiv:1804.10332},
  year={2018}
}

@article{kumar2021rma,
  title={Rma: Rapid motor adaptation for legged robots},
  author={Kumar, Ashish and Fu, Zipeng and Pathak, Deepak and Malik, Jitendra},
  journal={arXiv preprint arXiv:2107.04034},
  year={2021}
}

@article{ha2025learning,
  title={Learning-based legged locomotion: State of the art and future perspectives},
  author={Ha, Sehoon and Lee, Joonho and van de Panne, Michiel and Xie, Zhaoming and Yu, Wenhao and Khadiv, Majid},
  journal={The International Journal of Robotics Research},
  volume={44},
  number={8},
  pages={1396--1427},
  year={2025},
  publisher={SAGE Publications Sage UK: London, England}
}

@inproceedings{li2021reinforcement,
  title={Reinforcement learning for robust parameterized locomotion control of bipedal robots},
  author={Li, Zhongyu and Cheng, Xuxin and Peng, Xue Bin and Abbeel, Pieter and Levine, Sergey and Berseth, Glen and Sreenath, Koushil},
  booktitle={2021 IEEE International Conference on Robotics and Automation (ICRA)},
  pages={2811--2817},
  year={2021},
  organization={IEEE}
}

@article{margolis2024rapid,
  title={Rapid locomotion via reinforcement learning},
  author={Margolis, Gabriel B and Yang, Ge and Paigwar, Kartik and Chen, Tao and Agrawal, Pulkit},
  journal={The International Journal of Robotics Research},
  volume={43},
  number={4},
  pages={572--587},
  year={2024},
  publisher={SAGE Publications Sage UK: London, England}
}

@article{bussola2025guided,
  title={Guided Reinforcement Learning for Omnidirectional 3D Jumping in Quadruped Robots},
  author={Bussola, Riccardo and Focchi, Michele and Turrisi, Giulio and Semini, Claudio and Palopoli, Luigi},
  journal={arXiv preprint arXiv:2507.16481},
  year={2025}
}

@article{fahmi2022vital,
  title={Vital: Vision-based terrain-aware locomotion for legged robots},
  author={Fahmi, Shamel and Barasuol, Victor and Esteban, Domingo and Villarreal, Octavio and Semini, Claudio},
  journal={IEEE Transactions on Robotics},
  volume={39},
  number={2},
  pages={885--904},
  year={2022},
  publisher={IEEE}
}

@inproceedings{yu2021visual,
  title={Visual-locomotion: Learning to walk on complex terrains with vision},
  author={Yu, Wenhao and Jain, Deepali and Escontrela, Alejandro and Iscen, Atil and Xu, Peng and Coumans, Erwin and Ha, Sehoon and Tan, Jie and Zhang, Tingnan},
  booktitle={5th Annual Conference on Robot Learning},
  year={2021}
}

@inproceedings{duan2024learning,
  title={Learning vision-based bipedal locomotion for challenging terrain},
  author={Duan, Helei and Pandit, Bikram and Gadde, Mohitvishnu S and Van Marum, Bart and Dao, Jeremy and Kim, Chanho and Fern, Alan},
  booktitle={2024 IEEE International Conference on Robotics and Automation (ICRA)},
  pages={56--62},
  year={2024},
  organization={IEEE}
}

@inproceedings{jain2019hierarchical,
  title={Hierarchical reinforcement learning for quadruped locomotion},
  author={Jain, Deepali and Iscen, Atil and Caluwaerts, Ken},
  booktitle={2019 IEEE/RSJ international conference on intelligent robots and systems (IROS)},
  pages={7551--7557},
  year={2019},
  organization={IEEE}
}

@article{yang2021learning,
  title={Learning vision-guided quadrupedal locomotion end-to-end with cross-modal transformers},
  author={Yang, Ruihan and Zhang, Minghao and Hansen, Nicklas and Xu, Huazhe and Wang, Xiaolong},
  journal={arXiv preprint arXiv:2107.03996},
  year={2021}
}

@article{singh2022reinforcement,
  title={Reinforcement learning in robotic applications: a comprehensive survey},
  author={Singh, Bharat and Kumar, Rajesh and Singh, Vinay Pratap},
  journal={Artificial Intelligence Review},
  volume={55},
  number={2},
  pages={945--990},
  year={2022},
  publisher={Springer}
}

@inproceedings{xiao2024egocentric,
  title={Egocentric Visual Locomotion in a Quadruped Robot},
  author={Xiao, Feng and Chen, Ting and Li, Yong},
  booktitle={Proceedings of the 2024 8th International Conference on Electronic Information Technology and Computer Engineering},
  pages={172--177},
  year={2024}
}

@article{lai2024world,
  title={World model-based perception for visual legged locomotion},
  author={Lai, Hang and Cao, Jiahang and Xu, Jiafeng and Wu, Hongtao and Lin, Yunfeng and Kong, Tao and Yu, Yong and Zhang, Weinan},
  journal={arXiv preprint arXiv:2409.16784},
  year={2024}
}

@article{gu2021efficiently,
  title={Efficiently modeling long sequences with structured state spaces},
  author={Gu, Albert and Goel, Karan and R{\'e}, Christopher},
  journal={arXiv preprint arXiv:2111.00396},
  year={2021}
}

@article{gu2023mamba,
  title={Mamba: Linear-time sequence modeling with selective state spaces},
  author={Gu, Albert and Dao, Tri},
  journal={arXiv preprint arXiv:2312.00752},
  year={2023}
}

@inproceedings{li2025alignmamba,
  title={AlignMamba: Enhancing Multimodal Mamba with Local and Global Cross-modal Alignment},
  author={Li, Yan and Xing, Yifei and Lan, Xiangyuan and Li, Xin and Chen, Haifeng and Jiang, Dongmei},
  booktitle={Proceedings of the Computer Vision and Pattern Recognition Conference},
  pages={24774--24784},
  year={2025}
}

@article{liu2026cross,
  title={COMO: Cross-mamba interaction and offset-guided fusion for multimodal object detection},
  author={Liu, Chang and Ma, Xin and Yang, Xiaochen and Zhang, Yuxiang and Dong, Yanni},
  journal={Information Fusion},
  volume={125},
  pages={103414},
  year={2026},
  publisher={Elsevier}
}

@inproceedings{huang2024av,
  title={AV-Mamba: Cross-Modality Selective State Space Models for Audio-Visual Question Answering},
  author={Huang, Ziru and Li, Jia and Zhao, Wenjie and Guo, Yunhui and Tian, Yapeng},
  booktitle={Proceedings of the IEEE/CVF Conference on Computer Vision and Pattern Recognition Workshop (CVPRW)},
  pages={1--4},
  year={2024}
}

@article{xie2024fusionmamba,
  title={Fusionmamba: Dynamic feature enhancement for multimodal image fusion with mamba},
  author={Xie, Xinyu and Cui, Yawen and Tan, Tao and Zheng, Xubin and Yu, Zitong},
  journal={Visual Intelligence},
  volume={2},
  number={1},
  pages={37},
  year={2024},
  publisher={Springer}
}

@inproceedings{ye2025depmamba,
  title={Depmamba: Progressive fusion mamba for multimodal depression detection},
  author={Ye, Jiaxin and Zhang, Junping and Shan, Hongming},
  booktitle={ICASSP 2025-2025 IEEE International Conference on Acoustics, Speech and Signal Processing (ICASSP)},
  pages={1--5},
  year={2025},
  organization={IEEE}
}

@article{richard1957mark,
 ISSN = {00959057, 19435274},
 URL = {http://www.jstor.org/stable/24900506},
 author = {Richard Bellman},
 journal = {Journal of Mathematics and Mechanics},
 number = {5},
 pages = {679--684},
 publisher = {Indiana University Mathematics Department},
 title = {A Markovian Decision Process},
 urldate = {2025-08-16},
 volume = {6},
 year = {1957}
}

@article{kalman1960new,
  title={A new approach to linear filtering and prediction problems},
  author={Kalman, Rudolph Emil},
  year={1960}
}

@article{schulman2017proximal,
  title={Proximal policy optimization algorithms},
  author={Schulman, John and Wolski, Filip and Dhariwal, Prafulla and Radford, Alec and Klimov, Oleg},
  journal={arXiv preprint arXiv:1707.06347},
  year={2017}
}

@article{li2021survey,
  title={A survey of convolutional neural networks: analysis, applications, and prospects},
  author={Li, Zewen and Liu, Fan and Yang, Wenjie and Peng, Shouheng and Zhou, Jun},
  journal={IEEE transactions on neural networks and learning systems},
  volume={33},
  number={12},
  pages={6999--7019},
  year={2021},
  publisher={IEEE}
}

@article{coumans2021pybullet,
  title={PyBullet quickstart guide},
  author={Coumans, Erwin and Bai, Yunfei},
  journal={ed: PyBullet Quickstart Guide. https://docs. google. com/document/u/1/d},
  year={2021}
}

@article{tan2009computation,
  title={Computation of stabilizing PI-PD controllers},
  author={Tan, Nusret},
  journal={International Journal of Control, Automation and Systems},
  volume={7},
  number={2},
  pages={175--184},
  year={2009},
  publisher={Springer}
}

@inproceedings{mehta2020active,
  title={Active domain randomization},
  author={Mehta, Bhairav and Diaz, Manfred and Golemo, Florian and Pal, Christopher J and Paull, Liam},
  booktitle={Conference on Robot Learning},
  pages={1162--1176},
  year={2020},
  organization={PMLR}
}

@article{wang2021survey,
  title={A survey on curriculum learning},
  author={Wang, Xin and Chen, Yudong and Zhu, Wenwu},
  journal={IEEE transactions on pattern analysis and machine intelligence},
  volume={44},
  number={9},
  pages={4555--4576},
  year={2021},
  publisher={IEEE}
}


\end{document}